\RequirePackage{fix-cm}
\documentclass[twocolumn]{svjour3-arxiv}          

\smartqed  
\usepackage{graphicx}
\usepackage{bbold}
\usepackage{mathptmx}   
\usepackage{amsmath}  
\usepackage{paralist}
\usepackage{latexsym}
\usepackage{microtype}
\usepackage{booktabs}
\usepackage{multirow}
\usepackage{tabularx}
\usepackage[ruled,vlined]{algorithm2e}
\usepackage[normalem]{ulem}
\usepackage{flushend}

\usepackage{xcolor}

\newcommand{\strikethrough}[1]{}
\newcommand{\ashis}[1]{#1}

\begin{document}

\title{Attribute-based Regularization of \strikethrough{VAE}Latent Spaces \ashis{for Variational Auto-Encoders}}

\author{Ashis Pati         \and
        Alexander Lerch 
}


\institute{Ashis Pati \& Alexander Lerch\at
              Center for Music Technology,  \\
              Georgia Institute of Technology, \\
              Atlanta, USA \\
              \email{ashis.pati@gatech.edu} \& {alexander.lerch@gatech.edu} 
}

\date{}

\maketitle

\begin{abstract}
  Selective manipulation of data attributes using deep generative models is an active area of research. In this paper, we present a novel method to structure the latent space of a Variational Auto-Encoder (VAE) to encode different continuous-valued attributes explicitly. This is accomplished by using an attribute regularization loss which enforces a monotonic relationship between the attribute values and the latent code of the dimension along which the attribute is to be encoded. Consequently, post training, the model can be used to manipulate the attribute by simply changing the latent code of the corresponding regularized dimension. The results obtained from several quantitative and qualitative experiments show that the proposed method leads to disentangled and interpretable latent spaces which can be used to effectively manipulate a wide range of data attributes spanning image and symbolic music domains. 

\keywords{Representation Learning \and Latent Space Disentanglement \and Latent Space Regularization \and Generative Modeling}
\end{abstract}

\section{Introduction}
\label{sec:intro}
  Over the last few years, deep generative models have emerged as powerful tools to create and manipulate data. These models have been applied to tasks in several domains such as image generation \cite{razavi_generating_2019,van_den_oord_conditional_2016,gatys_image_2016,ledig_photo-realistic_2017}, text generation \cite{zhang_adversarial_2017,dai_transformer-xl_2019}, speech generation \cite{akuzawa_expressive_2018,hsu_learning_2017} and music creation \cite{huang_music_2018,pati_learning_2019,roberts_hierarchical_2018,roberts_learning_2018}. More recently, there has been considerable focus on improving the controllability of these models by enabling selective manipulation of data attributes such as changing the gender of the person in an image \cite{lample_fader_2017}, changing the speaking style in a given speech excerpt \cite{wang_style_2018}, or increasing the density of notes in a musical excerpt \cite{roberts_hierarchical_2018,hadjeres_glsr-vae_2017}. 

  \begin{figure}[t]
    \centering
    \includegraphics[width=\columnwidth]{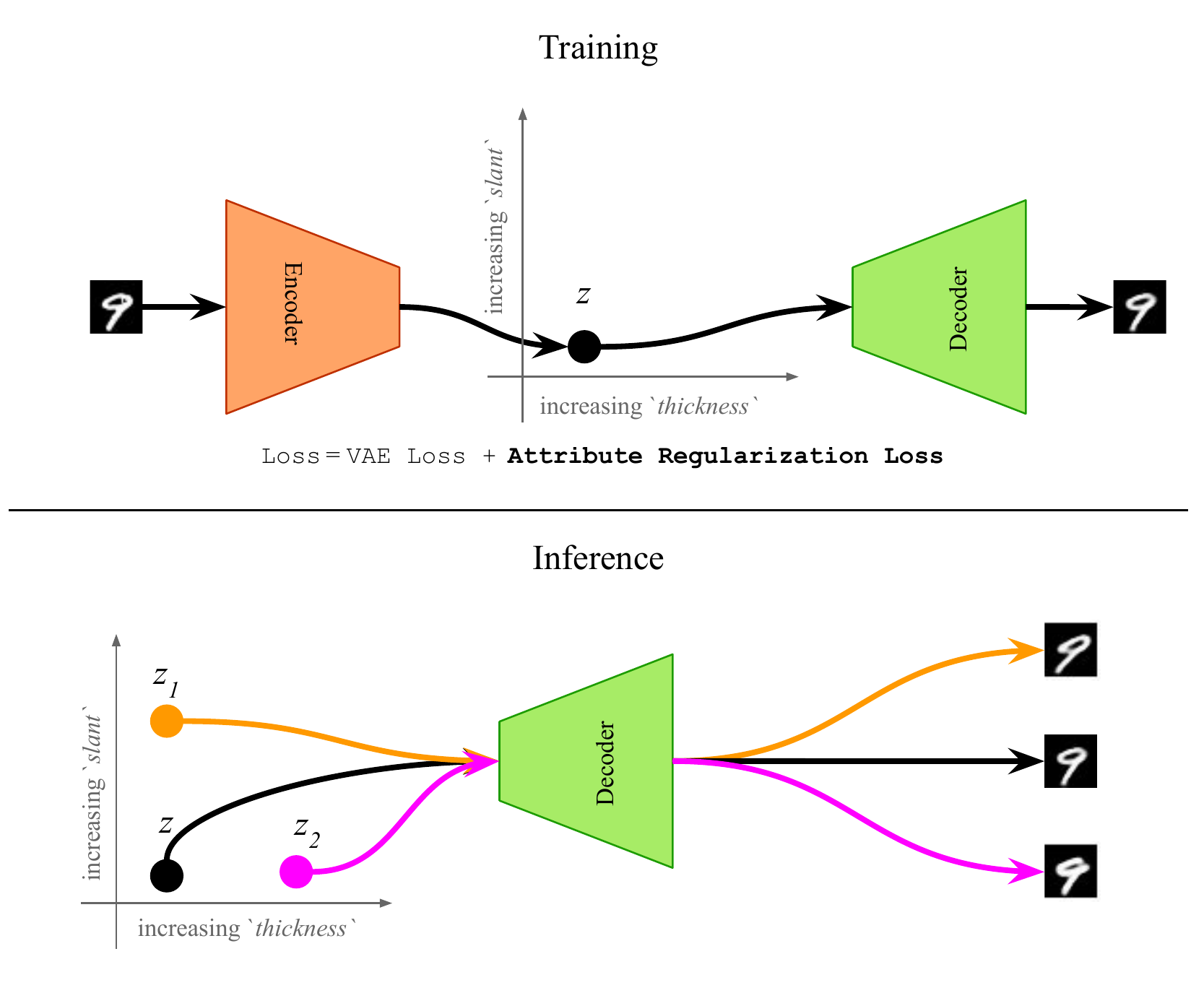}
    \caption{Motivation for the AR-VAE model which uses a novel attribute regularization loss (see Sect.~\ref{sec:ar_vae_loss}) during the training step to force the latent space to encode specific attributes along specific dimensions of the latent space of a VAE. During inference, individual data attributes can be manipulated by simply traversing along these regularized dimensions}
    \label{fig:1}     
  \end{figure}

  \begin{figure*}[t]
    \centering
    \begin{tabular}{@{}c@{}}
      \includegraphics[width=\columnwidth]{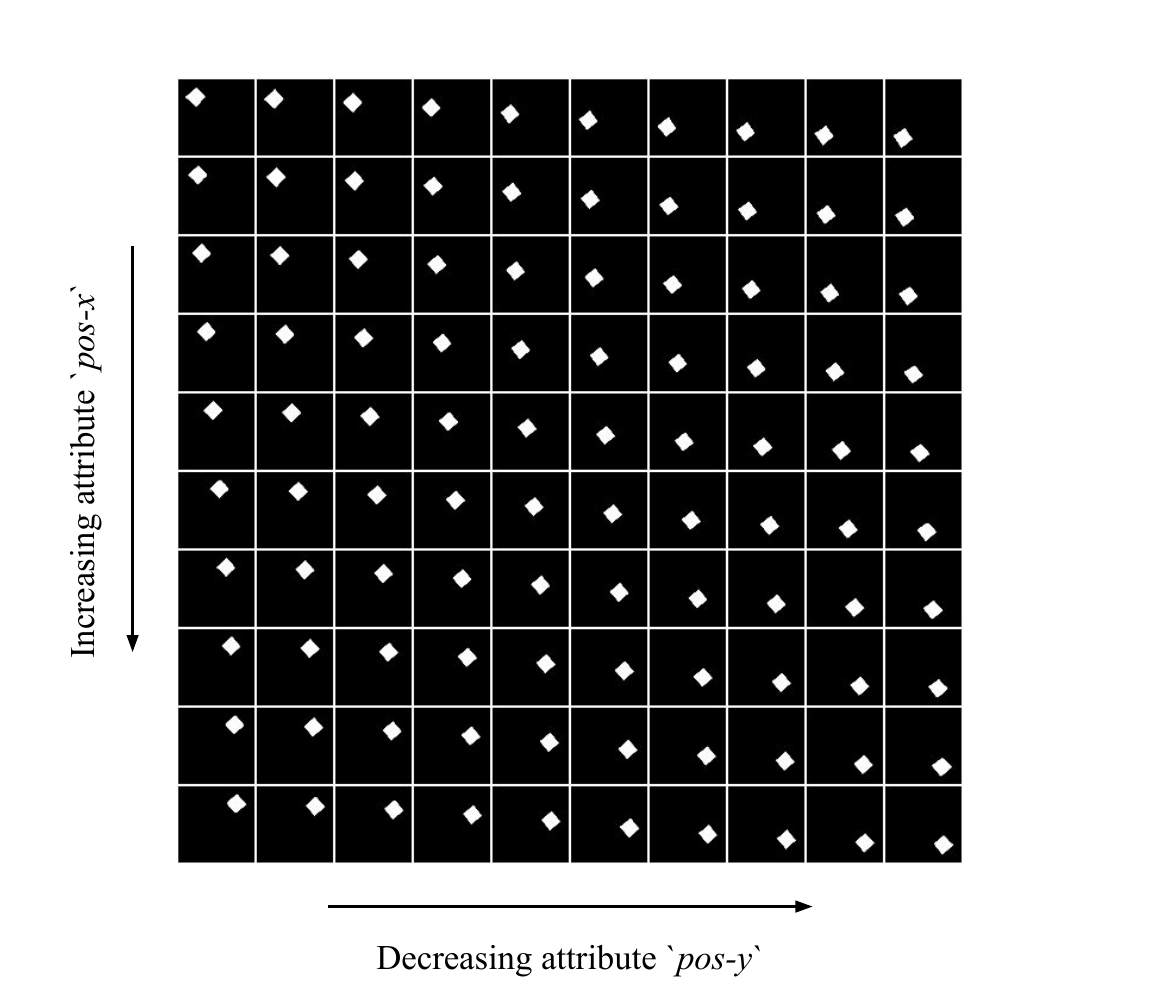} \\[\abovecaptionskip]
      \small (a)
    \end{tabular}
    \begin{tabular}{@{}c@{}}
      \includegraphics[width=\columnwidth]{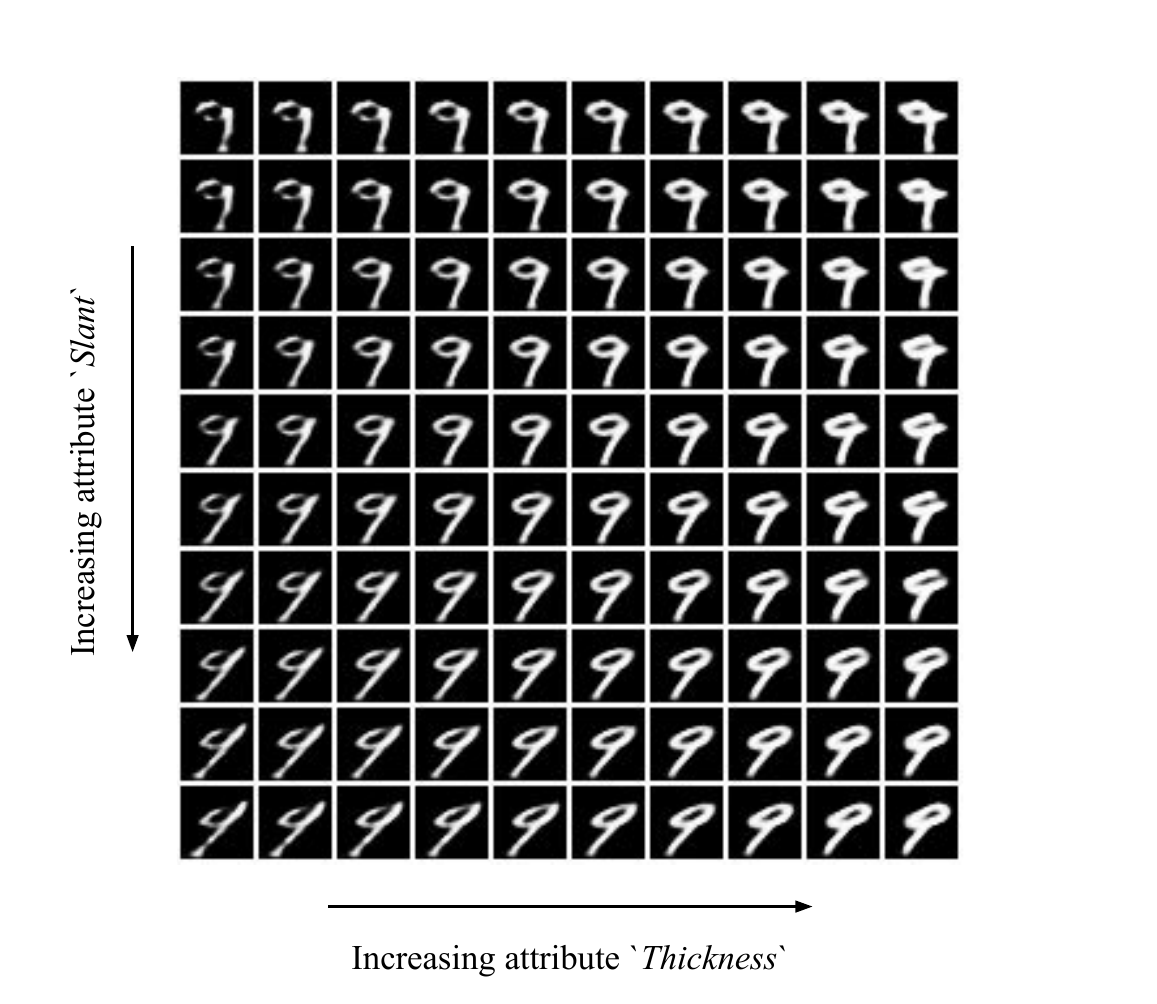} \\[\abovecaptionskip]
      \small (b)  
    \end{tabular}
    \caption{Manipulating two attributes independently by traversing along the corresponding regularized dimensions of the AR-VAE latent space. (a) 2-d sprites image \cite{matthey_dsprites_2017}, (b) Morpho-MNIST digit \cite{castro_morpho-mnist_2019} }
    \label{fig:2}
  \end{figure*}

  Latent representation-based models such as the Variational Auto-Encoders (VAEs) \cite{kingma_auto-encoding_2014} have shown promise in this direction as they are able to encode certain hidden attributes of the data \cite{carter_using_2017}. VAE-based latent spaces show interesting properties such as semantic interpolation \cite{roberts_learning_2018} and attribute vector arithmetic \cite{mikolov_distributed_2013,roberts_hierarchical_2018}. However, one of the key limitations of the vanilla-VAE framework is that the encoded attributes in the latent space cannot be explicitly controlled and the learnt attributes are often not interpretable by humans. This necessitates the need to rely on attribute vectors \cite{carter_using_2017}. In order to circumvent this limitation, there has been substantial research on modifying the VAE training procedure to learn representations which are able to disentangle different data attributes using either unsupervised methods \cite{higgins_beta-vae_2017,chen_isolating_2018,rubenstein_learning_2018,kumar_variational_2017} or supervised techniques \cite{lample_fader_2017,hadjeres_glsr-vae_2017,engel_latent_2017,bouchacourt_multi-level_2018,kulkarni_deep_2015}. However, there are certain limitations associated with both categories of methods which we discuss in Sect.~\ref{sec:relwork}.

  In this paper, we propose Attribute-Regularized VAE (AR-VAE) which uses a new supervised training method to create structured latent spaces where specific attributes are forced to be encoded along specific dimensions of the latent space. In order to achieve this, we formulate a novel regularization loss which forces each specific attribute of interest to have a monotonic relationship with the latent code of the dimension along which we want to encode the attribute (hereafter referred to as the \textit{regularized dimension}). Fig.~\ref{fig:1} demonstrates this overall idea. The proposed approach has the key advantages that it:
  \begin{enumerate}[(a)]
    \item has a simple formulation with few hyperparameters, 
    \item works with continuous data attributes, and 
    \item is  agnostic to how the attributes are computed/obtained.
  \end{enumerate}
  Apart from the above, the supervised nature of the proposed method eliminates the need for any post-training analysis to interpret the relationship between the latent code and the data attributes. 

  We show that using the proposed method we are able to learn disentangled latent representations useful for manipulating a wide range of attributes across two different domains: images and music. The superior performance of AR-VAE compared to the baseline models is demonstrated using several objective and subjective experiments. Fig.~\ref{fig:2} shows an example of the capability of our model to control a couple of attributes of 2-d sprites \cite{matthey_dsprites_2017} and Morpho-MNIST digits \cite{castro_morpho-mnist_2019}.
    
  The remainder of this paper is organized as follows: firstly, we present an overview of the related work in Sect.~\ref{sec:relwork} followed by the description of our method in Sect.~\ref{sec:method}. We then provide details of our experimental set-up and results in Sect.~\ref{sec:experimental_setup} and \ref{sec:results}, respectively. Finally, Sect.~\ref{sec:conclusion} concludes the paper and outlines avenues for future research.

\section{Related Work}
\label{sec:relwork}
  The approaches for attribute-based control over the output of generative models can be grouped into two broad categories. The first group of methods attempts to disentangle different factors of variation in the data \cite{bengio_representation_2013}. The majority of the methods in this category are unsupervised techniques. Approaches from the second category rely on identifying certain attributes of interest and using supervised techniques in order to enable control during the generation process.  

  \subsection{Unsupervised Disentanglement Learning}
  \label{sec:unsupervised}
    Unsupervised methods for disentanglement learning attempt to separate the distinct factors of variation in data \cite{bengio_representation_2013} and learn a representation where changes to a single underlying factor of variation leads to changes in a single factor of the learned representation \cite{locatello_challenging_2019}. Most of the current approaches to unsupervised disentanglement are based on variations of the VAE framework \cite{kingma_auto-encoding_2014} (see Sect.~\ref{sec:method_background} for a short introduction). The main idea behind these approaches is that forcing the latent representation to have a factorized aggregated posterior should result in disentanglement \cite{locatello_challenging_2019}. \ashis{This can be achieved using different means such as imposing constraints on the information capacity of the latent space \cite{higgins_beta-vae_2017,burgess_understanding_2018,rubenstein_learning_2018}, maximizing the mutual information between a subset of the latent code and the observations \cite{chen_infogan_2016}, and maximizing the independence between the latent variables \cite{chen_isolating_2018,kim_disentangling_2018}.}
    
    While many of these methods show good performance (based on one or more objective metrics for measuring disentanglement) on artificially generated image datasets (such as dSprites \cite{matthey_dsprites_2017}), a recent study by Locatello et al.\ shows that not only are these approaches sensitive to inductive biases such as choice of network, hyperparameters, and random seeds, but also that some amount of supervision is necessary for learning effective disentanglement \cite{locatello_challenging_2019}. In addition, since these methods seek to learn a factorized latent representation they work well for low-level data attributes. However, as we show in our experiments, they do not extend well for complex data attributes (which are usually some combination of low-level attributes). Consequently, their practical usefulness is limited. Using unsupervised methods for manipulating attributes also requires a post-training analysis to determine how different attributes are encoded along the different dimensions of the latent space.

  \subsection{Supervised Regularization Methods}
  \label{sec:supervised}
    There has also been some research on fully supervised methods to control attributes by learning the transformations \cite{yang_weakly-supervised_2015,reed_deep_2015}. However, these methods require availability of specifically annotated data regarding how each transformation changes a given data-point. Obtaining such data is costly and hence, such methods only have been shown to work with artificially created datasets. 
    
    An alternative is to use any available attribute information as conditioning inputs without needing explicit supervision. Methods within this group can broadly be differentiated into two categories,
    \begin{inparaenum}[(a)]
      \item methods relying on learning attribute-dependent latent spaces, and 
      \item methods relying on learning attribute-invariant latent spaces. 
    \end{inparaenum} 

    Attribute-dependent latent spaces attempt to explicitly encode different attributes along different dimensions of the latent space. These either encode individual attributes along individual dimensions \ashis{\cite{hadjeres_glsr-vae_2017,kulkarni_deep_2015},}\strikethrough{(e.g., GLSR-VAE, DC-IGN)} or decompose the latent space into different parts where each part corresponds to specific attributes \ashis{\cite{donahue_semantically_2018,brunner_midi-vae_2018}}
    \strikethrough{(e.g., the SD-GAN, MIDI-VAE)}. There have also been attempts to use latent spaces of vanilla-VAEs and learn transformation and traversal methods to improve interpretability \cite{adel_discovering_2018} and learn conditional generation \cite{engel_latent_2017,pati_learning_2019}. 

    Attribute-invariant latent spaces are those where the latent representation is independent of the attribute values. The idea is to learn a generalized latent representation which, when combined with the attribute specific information, generates a data-point with that attribute. Conditional VAEs and \ashis{Generative Adversarial Networks} (GANs) were early attempts at this \cite{sohn_learning_2015,yan_attribute2image_2016,mirza_conditional_2014}. The more recent Fader networks proposed by Lample et al.\ demonstrated the effectiveness of this idea for image generation by using an adversarial training scheme \cite{lample_fader_2017}.
    
    Supervised methods also have their limitations. On the one hand, some methods are constrained to work only with certain types of data attributes. For instance, the Fader network, although designed for categorical attributes, has been shown to only work well with binary attributes \cite{lample_fader_2017}. On the other hand, some methods might impose additional constraints. For example, \strikethrough{the DC-IGN requires}\ashis{they might require} the ability to generate or sample data-points by independently varying attributes \cite{kulkarni_deep_2015}, \ashis{or}\strikethrough{the ML-VAE and SD-GAN} require the ability to group data-points based on certain attributes \cite{bouchacourt_multi-level_2018,donahue_semantically_2018}. In addition, very few supervised approaches are designed to work with continuous-valued attributes. To our best knowledge, the only model which is designed to exclusively work with continuous-valued data attributes is the \ashis{Geodesic Latent Space Regularization VAE} (GLSR-VAE) \cite{hadjeres_glsr-vae_2017}. However, it requires differentiable computation of data attributes and careful tuning of hyperparameters. 

\section{Method}
\label{sec:method} 
  The goal of the proposed method is to train a structured latent space in which individual attributes are encoded along specific dimensions of the latent space. Such a latent structure enables controllable generation by selectively changing the latent code of the regularized dimension. For instance, if the attribute represents `thickness' of an MNIST digit, and the regularized dimension corresponds to the first dimension of the latent space, then sampling latent vectors with increasing values of the first dimension should result in digits with increasing thickness as illustrated in Fig.~\ref{fig:1}. We first present a brief background on VAEs before describing the regularization loss formulation and the learning algorithm for AR-VAE.

  \subsection{Background on Variational Auto-Encoders}
  \label{sec:method_background}
    A Variational Auto-Encoder (VAE)~\cite{kingma_auto-encoding_2014} is a generative model which uses an auto-encoding~\cite{vincent_extracting_2008} framework; during training, the model is forced to reconstruct its input. In a typical auto-encoder, the encoder learns to map data-points $\mathbf{x}$ from a high-dimensional data-space $X$ to points in a low-dimensional space $Z$. This low-dimensional space is referred to as the \textit{latent} space and points $\mathbf{z}$ in the latent space are called \textit{latent vectors}. The decoder learns to map the latent vectors back to the data-space. VAEs treat the latent vector as a random variable and model the generative process as a sequence of sampling operations: $\mathbf{z} \sim p(\mathbf{z})$, and $\mathbf{x} \sim p_{\theta}(\mathbf{x}|\mathbf{z})$, where $p_{\theta}(\mathbf{x}|\mathbf{z})$ is the decoder network parametrized by $\theta$, and $p(\mathbf{z})$ is a prior distribution over the latent space. The posterior $q_{\phi}(\mathbf{z}|\mathbf{x})$ is approximated by the encoder with parameters $\phi$. Variational Inference is used to approximate the posterior by minimizing the \ashis{Kullback-Leibler} (KL)-divergence~\cite{kullback_information_1951} between the approximate posterior and the true posterior by maximizing the evidence lower bound (ELBO)
    \begin{equation}\label{eq:VAE}
      \centering
      \log p(\mathbf{x}) \geq \mathbb{E}_{\mathbf{z} \sim q_{\phi}(\mathbf{z}|\mathbf{x})}[\log p_{\theta}(\mathbf{x}|\mathbf{z})] - \mathnormal{D}_{\mathrm{KL}}(q_{\phi}(\mathbf{z}|\mathbf{x})||p(\mathbf{z}))  
    \end{equation}
    where $\mathbb{E}[\cdot ]$ is the mathematical expectation, $\mathnormal{D}_{\mathrm{KL}}(\cdot || \cdot)$ is the KL-divergence.

    The first term of Eq.~(\ref{eq:VAE}) can be interpreted as maximizing the reconstruction accuracy while the second term ensures that realistic samples are generated when latent vectors are sampled using the prior $p(\mathbf{z})$~\cite{roberts_hierarchical_2018}. In practice, the VAE optimization minimizes the loss function
    \begin{equation}
      \mathnormal{L}_{\mathrm{VAE}}(\theta, \phi) = \mathnormal{L}_{\mathrm{recons}}(\theta, \phi) + \mathnormal{L}_{\mathrm{KLD}}(\theta, \phi)
    \end{equation}
    where $\mathnormal{L}_{\mathrm{recons}}(\theta, \phi)$ and $\mathnormal{L}_{\mathrm{KLD}}(\theta, \phi)$ are VAE reconstruction loss and the KL-Divergence regularization, respectively. The reconstruction loss is:
    \begin{equation}
      \label{eq:recons_loss}
      \mathnormal{L}_{\mathrm{recons}}(\theta, \phi) = \frac{1}{N} \sum_{i=1}^{N}\left \|  \hat{\mathbf{x}_{i}} - \mathbf{x}_{i}  \right \|_{2}^{2}
    \end{equation}
    where $N$ is the number of examples, $\hat{\mathbf{x}}$ is the reconstruction of $\mathbf{x}$ obtained using the encoder and decoder of the VAE. The L-2 norm in the above equation is replaced by cross-entropy loss for class prediction tasks. The regularization loss is
    \begin{equation}
      \label{eq:KLD_loss}
      \mathnormal{L}_{\mathrm{KLD}}(\theta, \phi) = \mathnormal{D}_{\mathrm{KL}}(q_{\phi}(\mathbf{z}|\mathbf{x})||p(\mathbf{z})).
    \end{equation}

    As mentioned in Sect.~\ref{sec:relwork}, most unsupervised methods for learning disentangled representations use a slight modification of this objective. For instance, the $\beta$-VAE model \cite{higgins_beta-vae_2017,bowman_generating_2016} uses the following formulation
    \begin{equation}
      \label{eq:loss_beta_vae}
      \mathnormal{L}_{\mathrm{VAE}}(\theta, \phi) = \mathnormal{L}_{\mathrm{recons}}(\theta, \phi) + \beta \mathnormal{L}_{\mathrm{KLD}}(\theta, \phi).
    \end{equation}
    The core idea here is that using $\beta > 1$ encourages the independence of the dimensions of the latent space and leads to better disentanglement. However, the trade off is that increasing $\beta$ might result in reduced reconstruction quality.

  \subsection{Attribute Regularization Loss}
  \label{sec:ar_vae_loss}
    Let us consider a $\mathbb{D}$-dimensional latent space where latent vectors are represented as $\mathbf{z}: \left \{ z^k \right \}, k \in [0,\mathbb{D})$. Formally, the objective is to encode an attribute $a$ along a dimension $r$ of the latent space such that, as we traverse along $r$, the attribute value $a$ of the generated data increases. Mathematically, if $a(\mathbf{x}_i)>a(\mathbf{x}_j)$, where $\mathbf{x}_i$ and $\mathbf{x}_j$ are two data-points generated using latent vectors $\mathbf{z}_i$ and $\mathbf{z}_j$, then $z^{r}_{i} > z^{r}_{j}$ should hold for any arbitrary $i$ and $j$. 

    This is accomplished by adding an attribute-specific regularization loss to the VAE training objective. To compute this loss, we use a mini-batch containing $m$ training examples and follow a three step process: 

    \begin{enumerate}
      \item An attribute distance matrix $\mathnormal{D}_a \in \mathbb{R}^{m \times m}$ is computed for all examples in the training mini-batch:
        \begin{equation}
          \mathnormal{D}_a(i,j) = a(\mathbf{x}_i) - a(\mathbf{x}_j)
        \end{equation}
      where $i,j \in [0,m)$.
      \item Next, a similar distance matrix $\mathnormal{D}_r \in \mathbb{R}^{m \times m}$ is computed for the regularized dimension $r$ of the latent vectors:
        \begin{equation}
          \mathnormal{D}_r(i,j) = z^{r}_{i} - z^{r}_{j}.
        \end{equation}
      \item The regularization loss is finally formulated as:
        \begin{equation}
          \label{eq:reg_loss}
          \mathnormal{L}_{r,a} = \mathrm{MAE}(\tanh \left ( \delta \mathnormal{D}_r \right ) - \mathrm{sgn} \left ( \mathnormal{D}_a \right ))  ,
        \end{equation}
      where $\mathrm{MAE}\left ( \cdot \right )$ is the mean absolute error, $\tanh \left ( \cdot \right )$ is the hyperbolic tangent function, $\mathrm{sgn} \left ( \cdot \right )$ is the sign function, and $\delta$ is a tunable hyperparameter which decides the spread of the posterior distribution.
    \end{enumerate}

    The $\mathrm{sgn} \left ( \cdot \right )$ function is used with the attribute distance matrix $\mathnormal{D}_a$ since we are only interested in whether the attribute value of a data-point is higher or lower than the others in the mini-batch and do not care about the magnitude of the differences. The $\tanh \left ( \cdot \right )$ function is used for the regularized dimension distance matrix $\mathnormal{D}_r$ since it has the same range as $\mathrm{sgn} \left ( \mathnormal{D}_a \right )$, i.e, $[-1, 1]$, and we want to minimize the mean absolute error between $\tanh \left ( \delta \mathnormal{D}_r \right )$ and $\mathrm{sgn} \left ( \mathnormal{D}_a \right )$. In addition, $\tanh \left ( \cdot \right )$ is a differentiable function which ensures that the loss is differentiable with respect to the latent vectors (and consequently the encoder parameters). Thus, we can use gradient descent for optimizing the network parameters. Other functions having such properties could also be potentially used to replace $\tanh \left ( \cdot \right )$.
    
    Overall, this formulation forces the latent code of the regularized dimension to have a monotonic relationship with the attribute values. Note that, unlike the GLSR-VAE formulation \cite{hadjeres_glsr-vae_2017} which requires that the attributes should be computed using differentiable functions, our formulation is agnostic to the way in which the attributes are computed/obtained.

\subsection{Learning Algorithm}
\label{sec:algo_learn}
  If our attribute set $\mathcal{A}: \left \{ a_{l} \right \}, l \in [0,\mathbb{L})$, contains $\mathbb{L}$ attributes, ($\mathbb{L} \leq \mathbb{D}$), then the overall loss function for AR-VAE is formulated as
  \begin{equation}
    \label{eq:loss_ar_vae}
    \mathnormal{L}_{\mathrm{AR\text{-}VAE}} = \mathnormal{L}_{\mathrm{recons}} + \beta \mathnormal{L}_{\mathrm{KLD}} + \gamma \sum_{l=0}^{\mathbb{L}-1}\mathnormal{L}_{r_{l},a_{l}}
  \end{equation}
  where $\mathnormal{L}_{r_{l},a_{l}}$ is the regularization loss for the attribute $a_{l}$ with $r_{l}$ as the index of the regularized dimension, and $\gamma$ is a tunable hyperparameter which we call as the regularization strength. We omit $\theta$ and $\phi$ in the above equation for brevity. The overall learning algorithm for AR-VAE is shown in Algorithm~\ref{algo1}. 
  
  \begin{algorithm}[h]
    \label{algo1}
    \SetAlgoLined
    \DontPrintSemicolon
    \SetKwInput{KwIn}{Input}
    \KwIn{observations and attribute labels $(\mathbf{x}_{i}, \left\{ a_{l} \right\}_{i})_{i=1}^{N} $ ($N$ is the number of examples in the dataset, $l \in [0, \mathbb{L})$ where $\mathbb{L}$ is the number of attributes), batch-size $m$, indices of the latent dimensions to be regularized $\left\{r_{l} \right\}_{l=0}^{\mathbb{L}-1}$, initialized VAE encoder and decoder parameters $\theta, \phi$, neural network optimizer $g$}
      \Repeat{convergence of objective}{
        Randomly sample a batch of $m$ data-points $(\mathbf{x}_{i}, \left\{ a_{l} \right\}_{i})$ \;
        Compute $L_{\mathrm{recons}}(\theta, \phi)$ using Eq.~(\ref{eq:recons_loss}) \;
        Compute $L_{\mathrm{KLD}}(\theta, \phi)$ using Eq.~(\ref{eq:KLD_loss}) \;
        \For{$l \in [0,\mathbb{L})$}{
          Compute $\mathnormal{L}_{r_{l},a_{l}}$ using Eq.~(\ref{eq:reg_loss}) \;
        }
        Compute $L_{\mathrm{AR-VAE}}(\theta, \phi)$ using Eq.~(\ref{eq:loss_ar_vae}) \;
        Update VAE encoder and decoder parameters: $\theta, \phi \leftarrow g(L_{\mathrm{AR-VAE}}(\theta, \phi))$ \;
      }
      \caption{Learning algorithm for AR-VAE}
  \end{algorithm}

\section{Experimental Setup}
\label{sec:experimental_setup}
  In order to demonstrate the capabilities of the AR-VAE model, several quantitative and qualitative experiments are conducted. The quantitative experiments comprise of evaluating the degree of disentanglement of the latent space, reconstruction fidelity, ability to preserve content, and sensitivity to the two hyperparameters. The qualitative experiments comprise of manipulating data attributes using latent interpolations and traversals and visualizing the latent space with respect to the different attributes. Before delving deeper into the experimental results, we first present a brief description of the different datasets and attributes used, the baseline considered and some of the details related to the implementation of the models.  

  \subsection{Datasets and Attributes}
    Most of the current research on disentanglement learning use image-based datasets such as 2-d sprites \cite{matthey_dsprites_2017}, 3-d shapes \cite{burges_3d-shapes_2020}, CelebA \cite{liu_deep_2015}, and 3-D chairs \cite{aubry_seeing_2014} for evaluation. This might be limiting since most methods are developed, tested and validated on a single data domain (images). In addition, some of these datasets such as 2-d sprites and 3-d shapes are artificially generated. While artificially created datasets are useful for benchmarking different methods, they have relatively simpler factors of variation. Considering this, we use datasets from two different domains (images and music), include both artificial and real-world data, and consider a diverse set of attributes.

    The first dataset from the image domain is the 2-d sprites dataset which contains ($\approx 0.7$ million) two dimensional shapes having 5 simple factors of variation: \textit{shape}, \textit{scale}, \textit{orientation}, \textit{x-position}, and \textit{y-position} \cite{matthey_dsprites_2017}. This is standard dataset for evaluating disentanglement methods. The second dataset is the Morho-MNIST dataset which contains the $70000$ handwritten MNIST digits along with complex morphological attributes for each digit obtained using computational methods \cite{castro_morpho-mnist_2019}. The attributes are: \textit{area}, \textit{length}, \textit{thickness}, \textit{slant}, \textit{width}, and \textit{height}.

    From the music domain, two datasets consisting of single measures (bars) of monophonic melodies are used. The first dataset consists of measures extracted from the soprano parts of the J.S.\ Bach Chorales dataset \cite{cuthbert_music21_2010} ($\approx 350$ chorales). The second dataset consists of measures extracted from ($\approx 20000$) folk melodies in the Scottish and Irish style \cite{sturm_music_2016}. Since pitch and rhythm are the two primary features of a monophonic melody, the following four attributes are considered for both datasets:
    \begin{inparaenum}[(a)]
      \item \textit{rhythmic complexity}, based on Toussaint’s metrical complexity measure \cite{toussaint_mathematical_2002},
      \item \textit{pitch range}, the difference of lowest pitch value (in MIDI) in the measure from the highest pitch value,
      \item \textit{note density}, the count of the number of notes per measure, and
      \item \textit{contour}, the degree to which the melody moves up or down measured by summing up the difference in pitch values of all the notes in the measure.
    \end{inparaenum}
    All these attributes are generated using computational methods (see Appendix~\ref{app:musical-metrics} for details).

  \subsection{Baseline}
    Most of the unsupervised methods for disentanglement learning perform at par with one another \cite{locatello_challenging_2019}. Since the AR-VAE is effectively an extension of the $\beta$-VAE \cite{higgins_beta-vae_2017} (compare Eq.~(\ref{eq:loss_beta_vae}) to Eq.~(\ref{eq:loss_ar_vae})), the latter is chosen as a baseline for comparison. Consequently, it also doubly serves as an ablation case. In addition, the $\beta$-VAE model is shown to perform at par with other \ashis{supervised} models \cite{higgins_beta-vae_2017}\strikethrough{such as Info-GAN  and (DC-IGN) \cite{}, with the later being a supervised method}.
    
    Other supervised methods such as GLSR-VAE \cite{hadjeres_glsr-vae_2017} and Fader networks \cite{lample_fader_2017} were also considered as potential baselines. However, the former requires differentiable computation of attributes (which cannot to applied to the datasets and attributes considered in this paper) and the later was designed for binary/categorical attributes. We investigated adapting the Fader network design to work with continuous attributes, however, the attempt did not lead to good results.

  \subsection{Implementation Details}
    Different model architectures are chosen for the different data domains. Convolutional architectures are used for the VAEs trained on images whereas recurrent architectures are used with the music-based datasets. To ensure consistency, all models for a particular dataset are trained for the same number of epochs using the same optimizer and learning rate. The model architectures and other details are provided in Appendix~\ref{app:model-arch} and in our GitHub repository.\footnote{https://github.com/ashispati/ar-vae}. 
    
    For each dataset, both the $\beta$-VAE and the AR-VAE models were trained with $10$ different random initializations. Based on initial experiments, for the $\beta$-VAE models, $\beta$ was chosen as $4.0$ and $0.001$ for the image and music datasets, respectively (the low value of $\beta$ was necessary to train the music VAE models \cite{pati_learning_2019,roberts_hierarchical_2018}). For AR-VAE, the models for the image datasets were trained with $\gamma=10.0$ and $\delta=1.0$, whereas those for the music datasets were trained with $\gamma=1.0$ and $\delta=10.0$.

    \begin{figure*}[t]
      \centering
      \begin{tabular}{@{}c@{}}
        \includegraphics[width=0.32\textwidth]{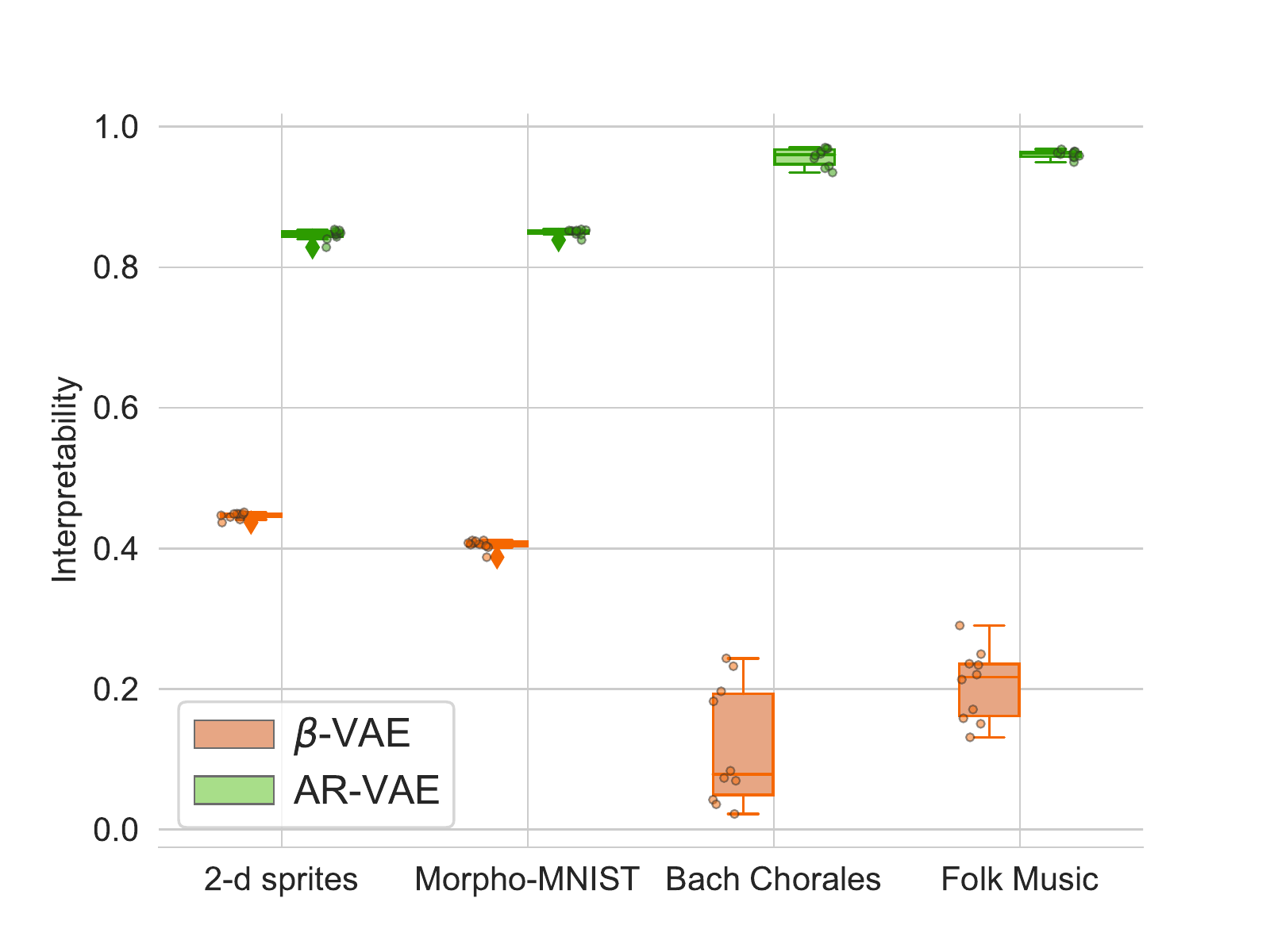} \\[\abovecaptionskip]
        \small (a) \textit{Interpretability}
      \end{tabular}
      \begin{tabular}{@{}c@{}}
        \includegraphics[width=0.32\textwidth]{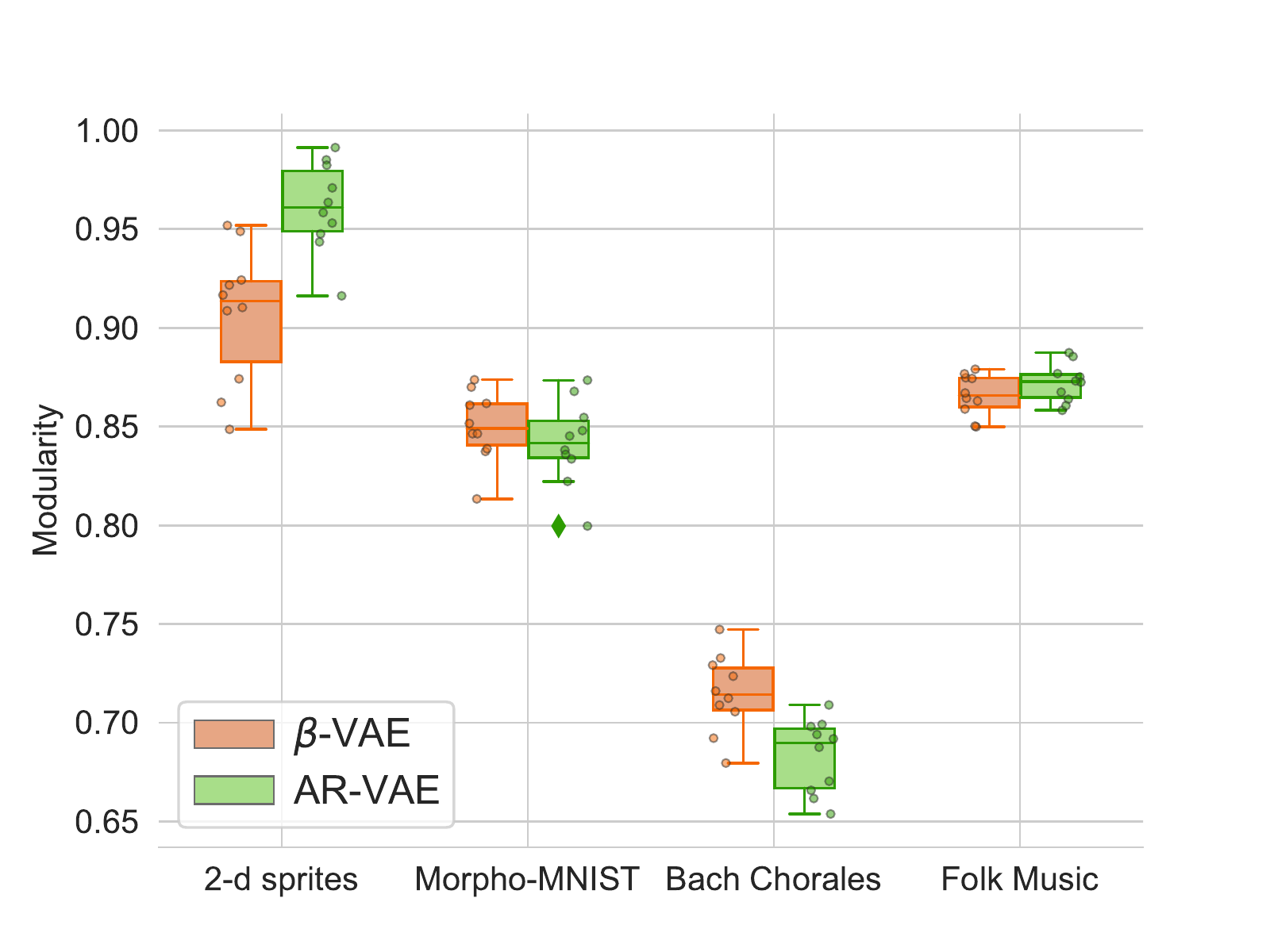} \\[\abovecaptionskip]
        \small (b) \textit{Modularity}
      \end{tabular}
      \begin{tabular}{@{}c@{}}
        \includegraphics[width=0.32\textwidth]{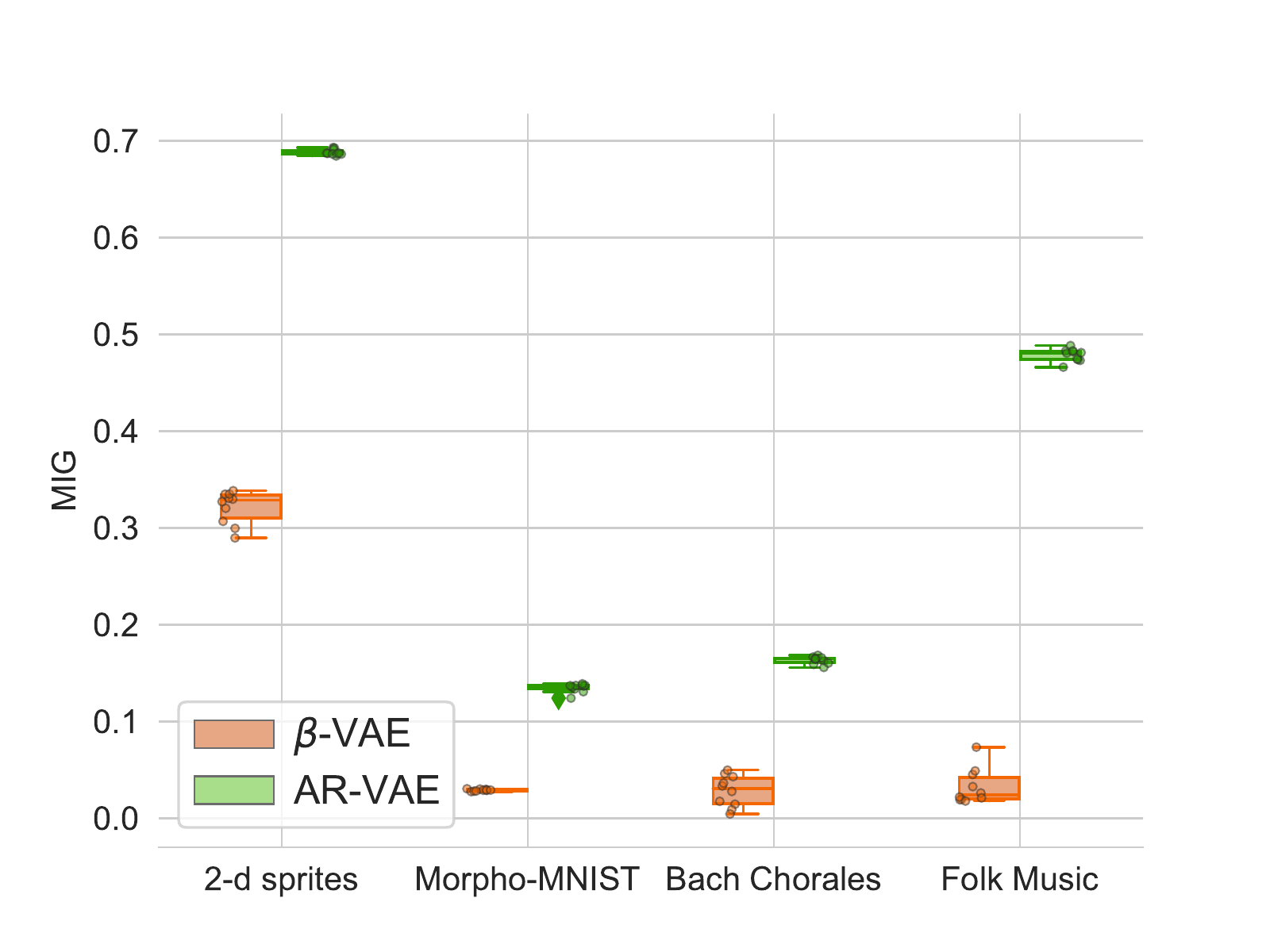} \\[\abovecaptionskip]
        \small (c) \textit{MIG}
      \end{tabular}
      \begin{tabular}{@{}c@{}}
        \includegraphics[width=0.32\textwidth]{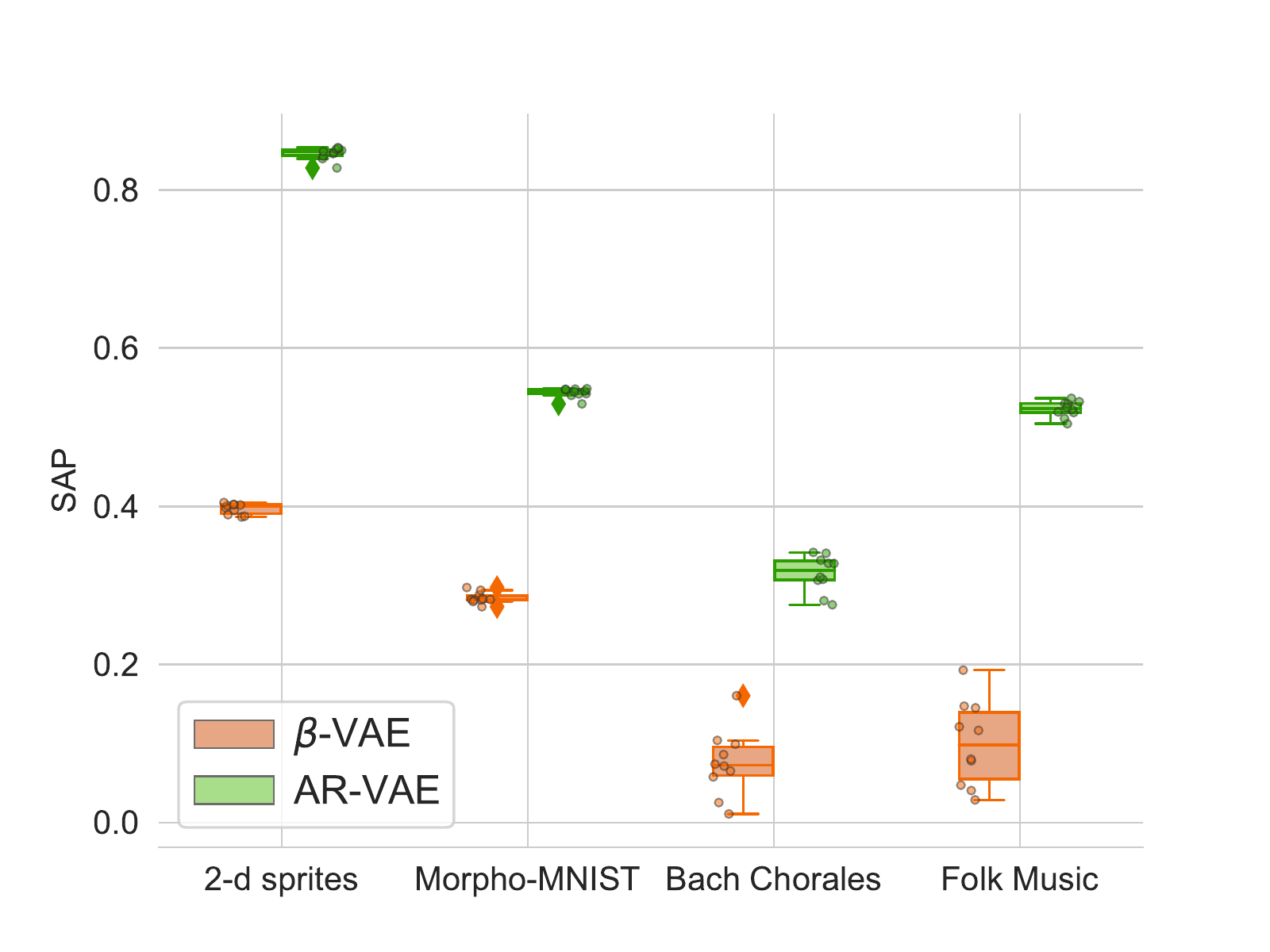} \\[\abovecaptionskip]
        \small (d) \textit{SAP}
      \end{tabular}
      \begin{tabular}{@{}c@{}}
        \includegraphics[width=0.32\textwidth]{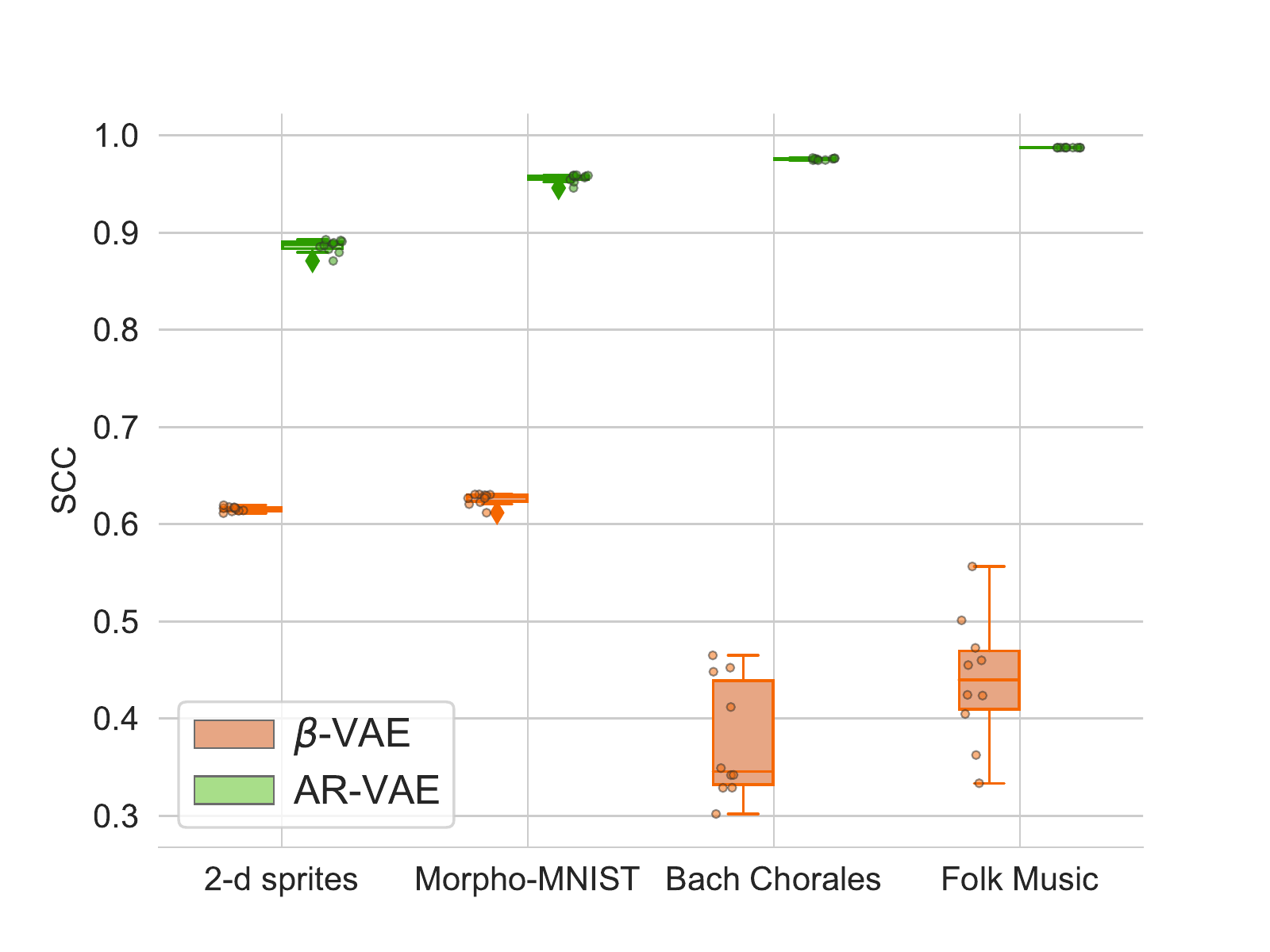} \\[\abovecaptionskip]
        \small (e) \textit{SCC}
      \end{tabular}
      \caption{Box plots for various disentanglement metrics (higher is better) across different datasets for ten different random seeds. The circular dots denote results for each random seed. AR-VAE scores better than $\beta$-VAE across the different datasets for all metrics except \textit{Modularity}} 
      \label{fig:dis_exp}
    \end{figure*}

\section{Results}
\label{sec:results}
  \subsection{Disentanglement}
  \label{exp_disentanglement}
  
    \begin{figure*}[t]
      \centering
      \includegraphics[width=0.8\textwidth]{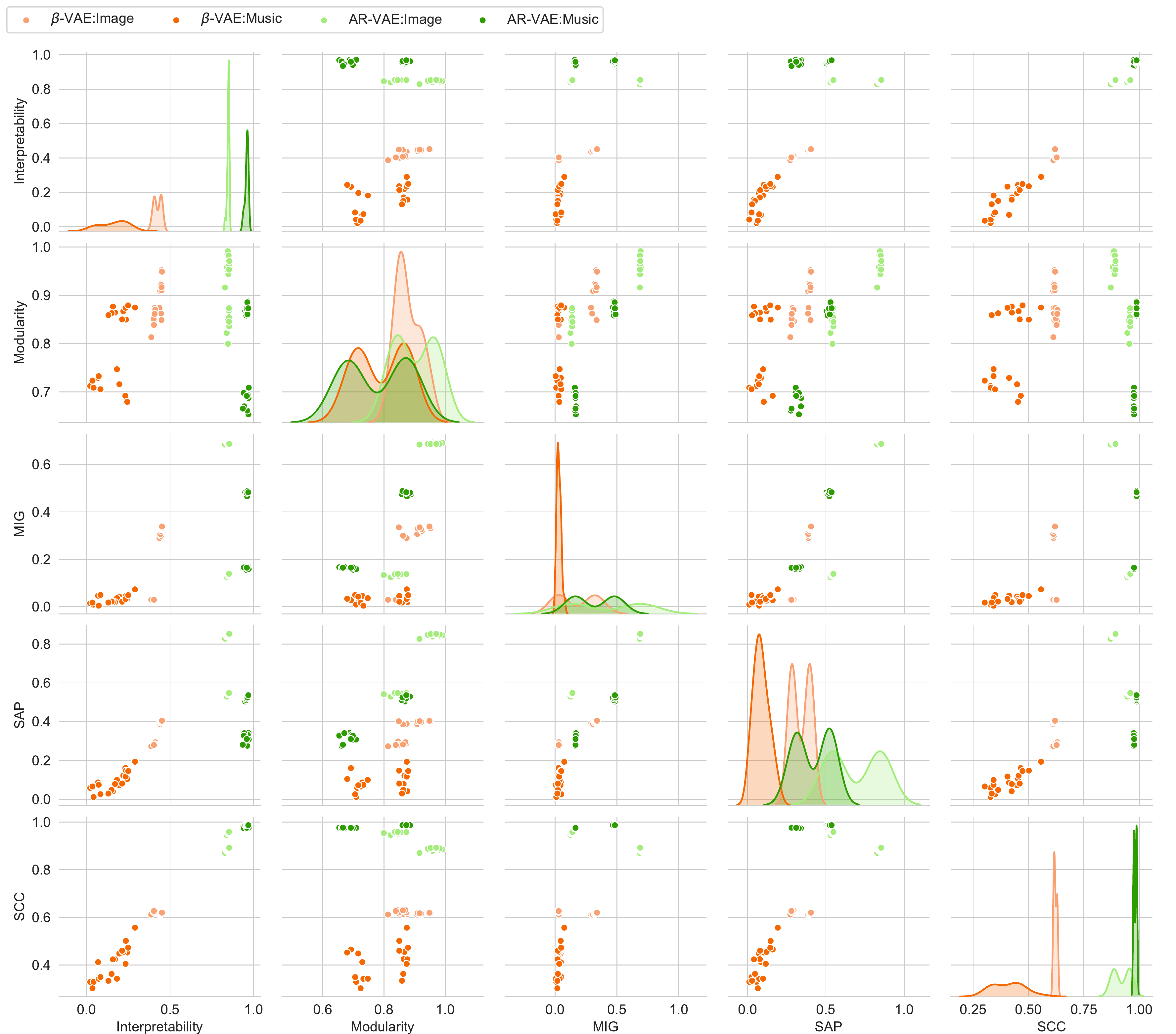}
      \caption{Pairwise scatter plots of the different metrics aggregated over the image and music datasets. For \textit{Interpretability} metric and \textit{SCC} score, the performance of the AR-VAE model is clearly separable from that of $\beta$-VAE even across different data domains. There is some overlap for other metrics, with \textit{Modularity} having the highest overlap}
      \label{fig:pair_plot}     
    \end{figure*}

    This experiment looks to evaluate the degree of disentanglement of the latent space with respect to the different data attributes. 
    
    There has been considerable work in the recent past to define objective metrics for measuring disentanglement of latent spaces \cite{adel_discovering_2018,ridgeway_learning_2018,chen_isolating_2018,kumar_variational_2017,higgins_beta-vae_2017,kim_disentangling_2018,eastwood_framework_2018}. Apart from disentangling the latent space, we are also interested in the degree to which AR-VAE is able to enforce a monotonic relationship between a given attribute and the regularized dimension. Taking this into account, the following metrics are used for this experiment:
    \begin{inparaenum}[(a)]
      \item \textit{Interpretability} \cite{adel_discovering_2018}, which measures the ability to predict a given attribute using only one dimension of the latent space, 
      \item \textit{Modularity} \cite{ridgeway_learning_2018}, which measures if each dimension of the latent space depends on only one attribute,
      \item \textit{Mutual Information Gap (MIG)} \cite{chen_isolating_2018}, which measures the difference of mutual information between a given attribute and the top two dimensions of the latent space which share maximum mutual information with the attribute,  
      \item \textit{Separated Attribute Predictability (SAP)}  \cite{kumar_variational_2017}, which measures the difference in the prediction error of the two most predictive dimensions of the latent space for a given attribute, and
      \item \textit{Spearman Correlation Coefficient (SCC)} score, which computes maximum value of the Spearman's correlation coefficient between an attribute and each dimension of the latent space. 
    \end{inparaenum}
    All metrics are aggregated by computing the mean across all attributes.

    Fig.~\ref{fig:dis_exp} shows the box plots with the performance of the models across all datasets and metrics. The following observations can be made. 
    Firstly, AR-VAE clearly outperforms $\beta$-VAE across all metrics (except \textit{Modularity}). Furthermore, this superior performance extends across all datasets in spite of the different domains and the varying degree of complexity in the attributes. The relatively poor performance for the \textit{Modularity} metric can be explained by looking at the normalization method used: the score is divided by the maximum value of the mutual information of all the attributes with a latent dimension. This can result in high \textit{Modularity} scores even with relatively low values of mutual information.
    Secondly, the highest improvement is seen for the \textit{Interpretability} metric and the \textit{SCC} score. This is also expected since AR-VAE forces a monotonic relationship between a given attribute and the latent code of the regularized dimension.
    Thirdly, there is lower improvement in \textit{MIG} which would suggest that there are other dimensions (apart from the regularized dimension) which share high mutual information with different attributes. This would also explain why the \textit{SAP} score does not improve as much as the \textit{Interpretability} metric. 
    Finally, for the image-based datasets, the metrics for 2-d sprites are generally higher when compared to the morpho-MNIST dataset. This could be due to the artificial nature of the former and it's simpler attributes. For the music-based datasets, Folk music has better performance than Bach chorales. This might be due to the significantly larger size of the Folk dataset.
    
    The superior performance of AR-VAE is also seen in the pairwise scatter plots shown in Fig.~\ref{fig:pair_plot} where the results of datasets belonging to the same domain are aggregated. There is a clear difference in the performance of the two models when either the \textit{Interpretability} metric or the \textit{SCC} score are considered. This is evident from the separation in the distributions for the two models in the diagonal plots for these metrics. The distributions for the \textit{MIG} and \textit{SAP} score show inter-domain separation although there seems to some overlap for datasets within each domain.

  \subsection{Reconstruction Fidelity}
  \label{exp_recons}
    The reconstruction quality is another important criterion. While a high degree of disentanglement is desirable, it should not come at the cost of reduced quality of the generation. 
    
\begin{figure}[t]
      \centering
      \includegraphics[width=0.35\textwidth]{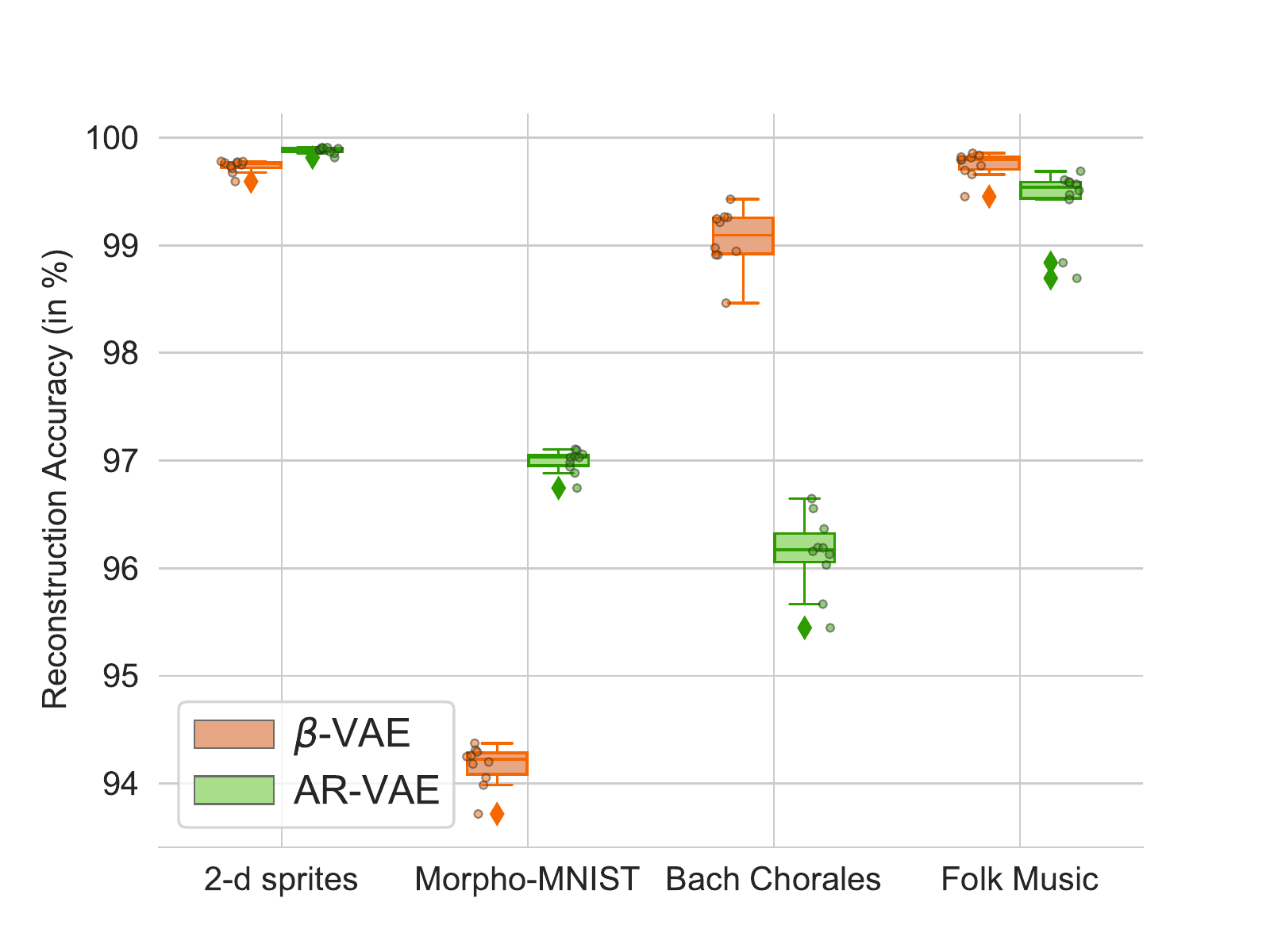}
      \caption{Box plots of reconstruction accuracies (higher is better) for different datasets. AR-VAE and $\beta$-VAE shows better performance in image and music datasets respectively\strikethrough{.}}
      \label{fig:recons_chart}     
    \end{figure}

\begin{figure}[t]
      \centering
      \includegraphics[width=\columnwidth]{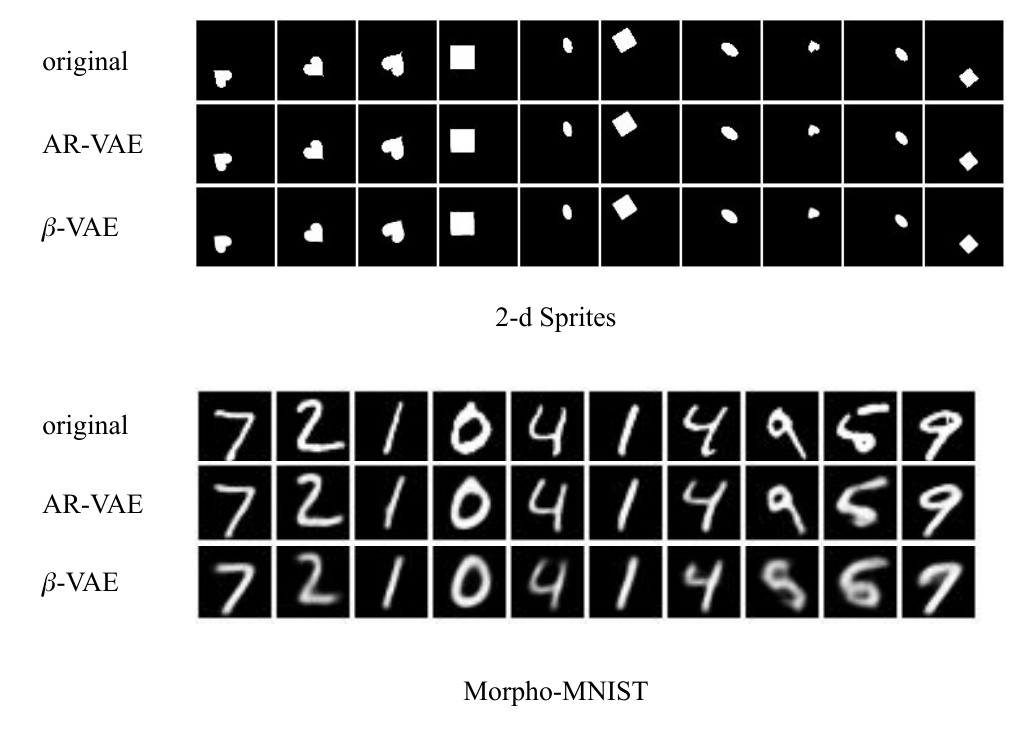}
      \caption{Few examples of reconstructions in the 2-d sprites and Morpho-MNIST datasets. The AR-VAE model seems to have better reconstructions compared to the $\beta$-VAE model}
      \label{fig:recons_examples}     
    \end{figure}

    Fig.~\ref{fig:recons_chart} shows the box plots for the reconstruction accuracies. While the performance of the two models is in the same range for the 2-d sprites dataset, AR-VAE performs better for the Morpho-MNIST dataset. A few example reconstructions for the 2-d sprites and Morpho-MNIST datasets are shown in Fig.~\ref{fig:recons_examples}. The better performance of AR-VAE can be gauged from are sharper reconstructions of MNIST digits. 

    For the music datasets, there is drop in performance for AR-VAE. The slight drop in case of the Folk dataset is expected since the same value of $\beta$ is used for both AR-VAE and $\beta$-VAE models. AR-VAE thus has more constraints than $\beta$-VAE during training. The larger drop in performance for the Bach Chorales might be due to the smaller size of the dataset. 

  \subsection{Content Preservation}
  \label{exp_content}
    \begin{figure}[t]
      \centering
      \includegraphics[width=0.35\textwidth]{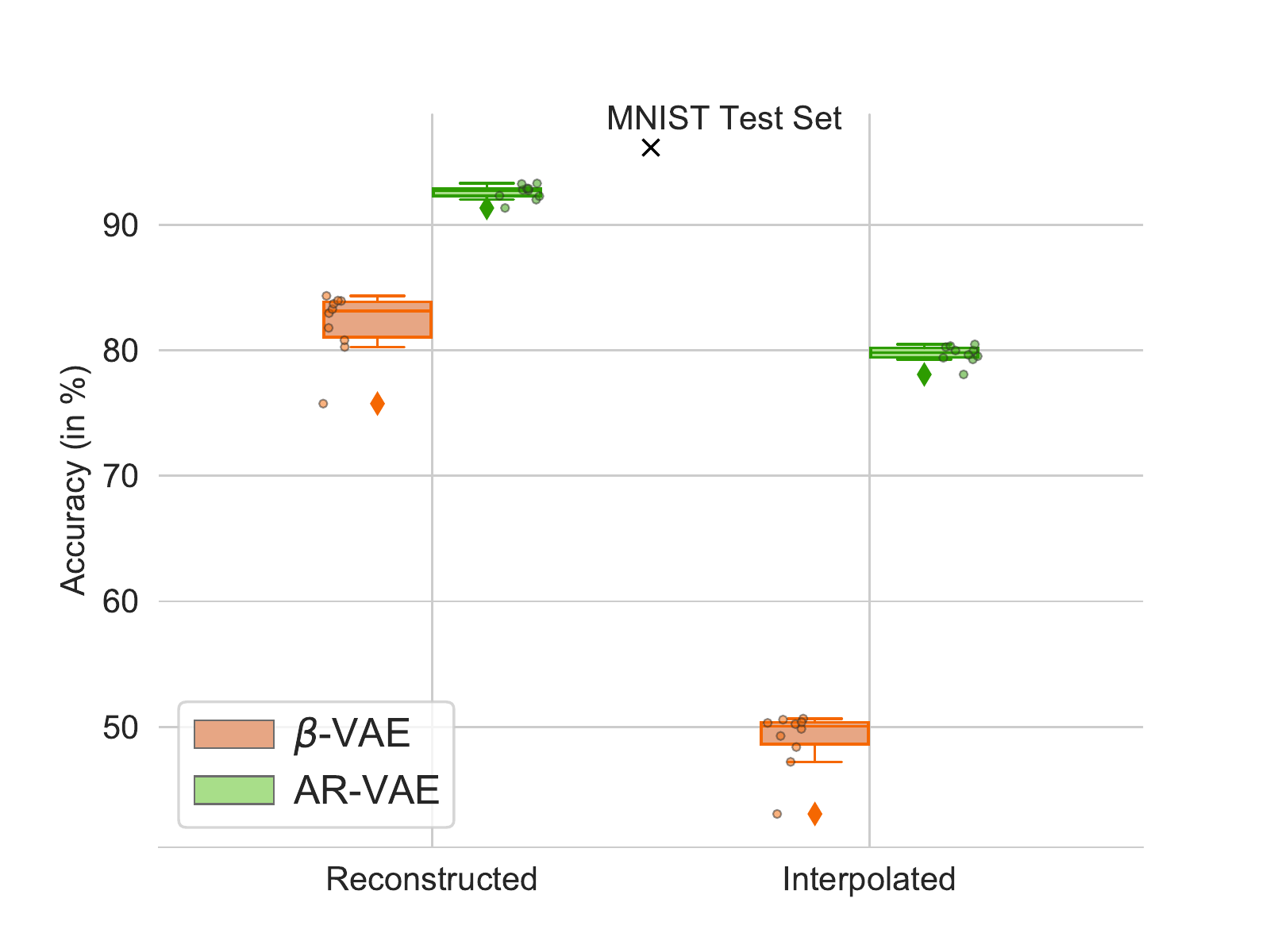}
      \caption{Distribution of accuracy (higher is better) in predicting the MNIST test set digits. Higher values indicate greater ability of the model in preserving the image content. `Reconstructed': prediction accuracy on reconstructed images from the MNIST test set, `Interpolated': prediction accuracy on interpolations generated by traversing along the regularized dimensions for different attributes. The prediction accuracy on the original MNIST test set is shown with a $\times$}
      \label{fig:digit_pred_acc}     
    \end{figure}
    In order to ascertain AR-VAE's ability to retain the content during interpolation, we conduct an additional experiment using the morpho-MNIST dataset using the identity of the digit as a proxy to the image content. Different variations of an input digit are generated by changing the attributes (via appropriate manipulation of the latent code) and then a pre-trained model is used to predict the digit class. The higher the classification accuracy, the better the model is at preserving the identity of the input. The experiment is run using the $10000$ digits in the MNIST test set. For each digit, $60$ variations are generated by interpolating along the regularized dimensions ($10$ for each of the six attributes). The pre-trained model has a ResNet-based architecture with an accuracy of $96.15\%$ on the unmodified MNIST test set. The results of this experiment are shown in Fig.~\ref{fig:digit_pred_acc}. While there is a only small difference in the performance of the two models in retaining the identity of the reconstructed digits, the AR-VAE model significantly outperforms the $\beta$-VAE for the interpolations. This indicates that AR-VAE is better for manipulating attributes while retaining the underlying content.

  \subsection{Hyperparameter Sensitivity}
  \label{exp_hyper_param}
    \begin{figure}[t]
      \centering
      \includegraphics[width=0.35\textwidth]{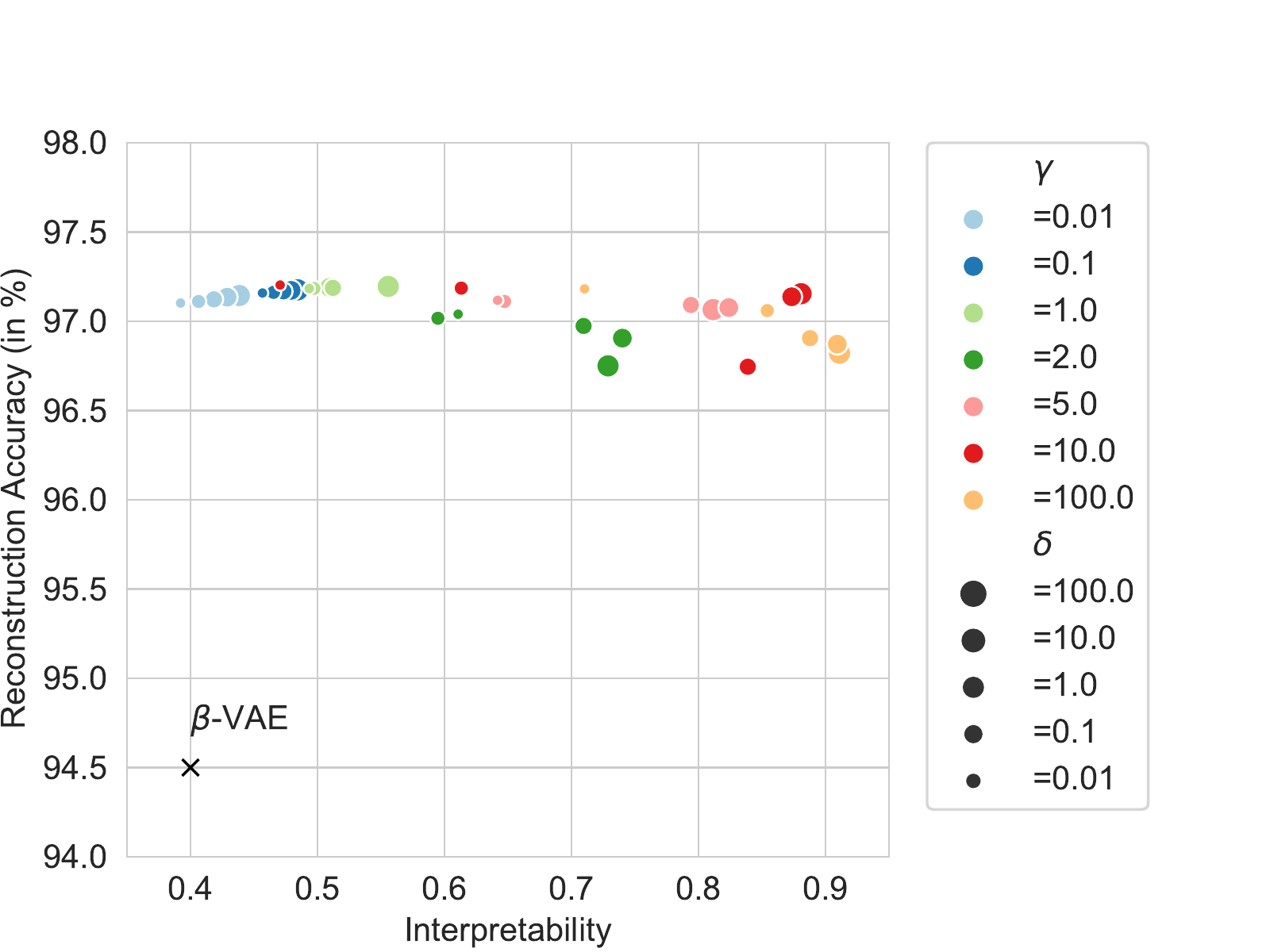}
      \caption{Effect of the hyperparameters $\gamma$ and $\delta$ on the performance of the AR-VAE model ($\beta$ is fixed at $1.0$) on the Morpho-MNIST dataset. Each dot corresponds to a unique combination of $\gamma$ and $\delta$ which are indicated by the color and the size of the dots respectively. The $\beta$-VAE model ($\beta=4$) is also shown (with a $\times$) for reference. The ideal model should lie on the top right corner of the plot (with high values of both reconstruction accuracy and \textit{Interpretability} metric)}
      \label{fig:hyper_param}     
    \end{figure}
    The next experiment assesses the sensitivity of the proposed model to the different hyperparameters ($\gamma$ and $\delta$). Fig.~\ref{fig:hyper_param} shows the trade-off between reconstruction accuracy and the \textit{Interpretability} metric as the hyperparameters are varied for the morpho-MNIST dataset while keeping $\beta$ fixed at $1.0$. 
    
    For lower values of $\gamma$, an increase in $\delta$ results in only marginal improvements in the \textit{Interpretability} metric without any loss in reconstruction accuracy. However, after $\gamma$ crosses $1.0$, increasing $\delta$ leads to significant improvement in the \textit{Interpretability} metric accompanied by a slight drop in the reconstruction accuracy. Note that the $\beta$-VAE model performs considerably worse in comparison.  Choosing $\gamma$ in the $[5.0,10.0]$ range and $\delta$ in the $[1.0, 10.0]$ range seems to give the best results. 

  \subsection{Attribute Manipulation}
  \label{exp_manipulation}
    This experiment presents a qualitative evaluation of the degree to which AR-VAE provides control over individual data attributes during the generation process. Given a data-point with latent code $z$, different variations for an attribute $a_{l}$ are generated by changing the latent code of the corresponding regularized dimension $z^{r_{l}}$ from $-4$ to $4$. For comparison, similar variations are generated for the $\beta$-VAE model by using the dimension which has the highest mutual information with the attribute. 

    \begin{figure}[t]
      \centering
      \includegraphics[width=\columnwidth]{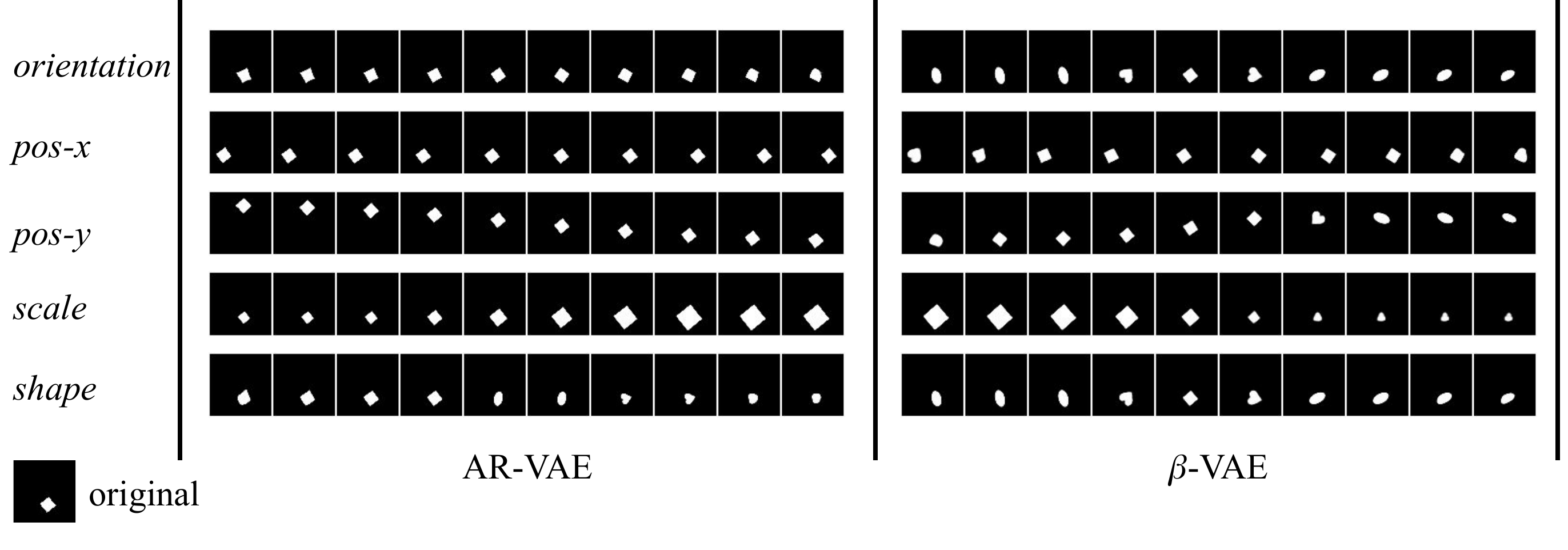}
      \caption{Controlling different attributes of a 2-d sprites image (shown in bottom left). AR-VAE model is able to manipulate each attribute independently. However, while the $\beta$-VAE model is also able to separately control \textit{pos-x}, \textit{pos-y}, and \textit{scale}, it changes the \textit{shape} of the sprite while manipulating \textit{orientation} and vice versa}
      \label{fig:interp_dsprites}     
    \end{figure}

    \begin{figure}[t]
      \centering
      \includegraphics[width=\columnwidth]{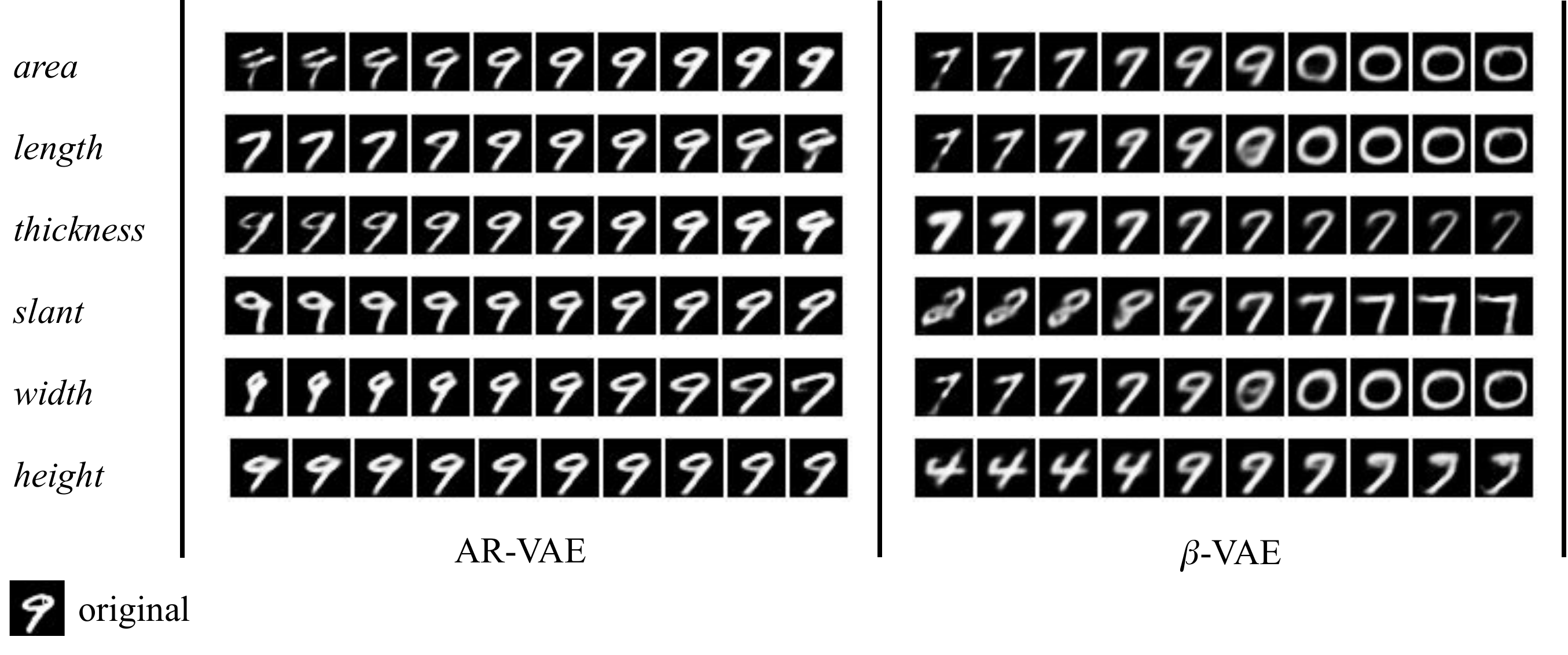}
      \caption{Controlling different attributes of an MNIST digit (shown in bottom left). The AR-VAE model is able to separately control the individual attributes of the original image and retains the identity of the digit in most cases. The $\beta$-VAE model, on the other hand, fails to retain the identity of original digit and the interpolations are not meaningful}
      \label{fig:interp_mnist}     
    \end{figure}

    \paragraph{Images:}  Fig.~\ref{fig:interp_dsprites} and~\ref{fig:interp_mnist} show the results of controlling attributes for the 2-d sprites and the Morpho-MNIST datasets, respectively. 
    While both models are able to control the individual attributes to a similar degree for the 2-d sprites dataset (Fig.~\ref{fig:interp_dsprites}), the AR-VAE model performs better for the Morpho-MNIST dataset (Fig.~\ref{fig:interp_mnist}). Not only are the interpolations meaningful with respect to the attribute of interest, but the AR-VAE is able to retain the identity of the original digit in most cases while $\beta$-VAE fails to do so (Sect.~\ref{exp_content} further supports this observation). See Appendix~\ref{app:add_results} for more examples.

    \begin{figure*}[t]
      \centering
      \begin{tabular}{@{}c@{}}
        \includegraphics[width=\textwidth]{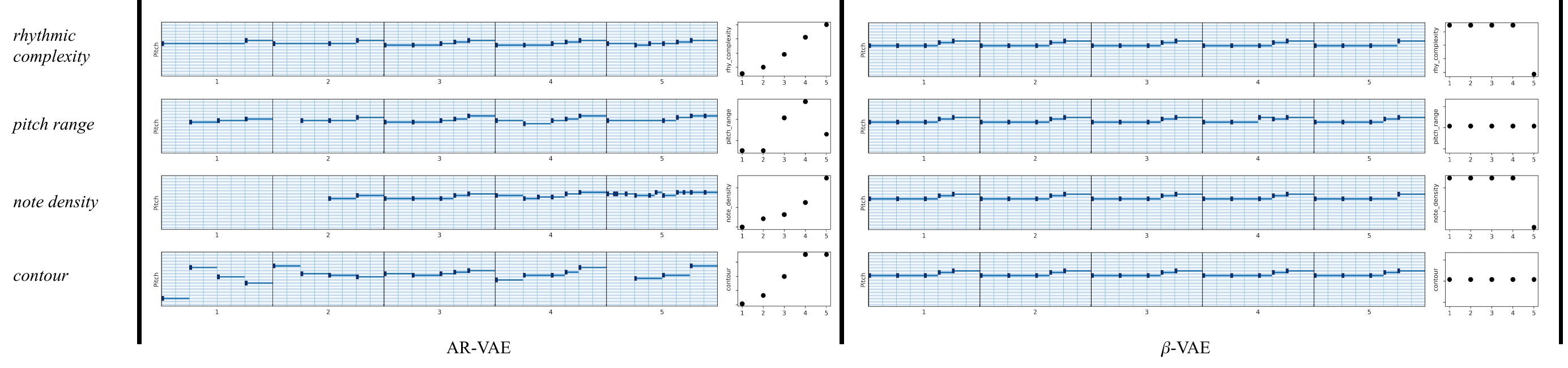} \\[\abovecaptionskip]
        \small (a) Bach Chorales
      \end{tabular}
      \begin{tabular}{@{}c@{}}
        \includegraphics[width=\textwidth]{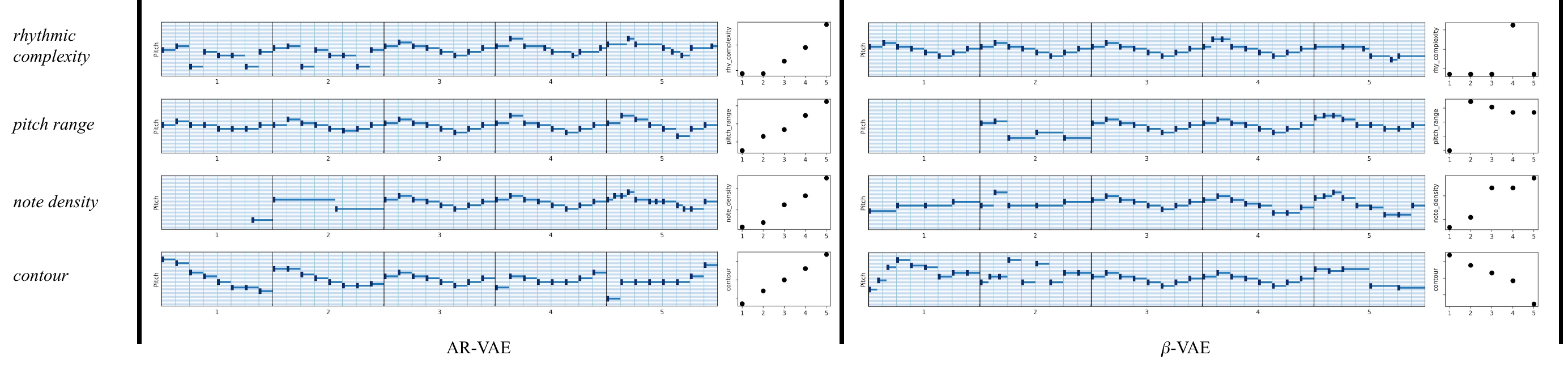} \\[\abovecaptionskip]
        \small (b) Folk Music
      \end{tabular}
      \caption{Controlling different attributes of musical measures. The piano-rolls show measures generated by increasing latent code of the regularized dimension (or in the case of $\beta$-VAE, the dimension with the highest mutual information) for the respective attribute. The light vertical lines within each measure denote the location of the eighth-notes. The \textit{y}-axis of the piano rolls show pitch in semi-tones. The plots on the right show how the attribute values change with the increase in the latent code. For AR-VAE, in most cases, traversing along the regularized dimension leads to increase in attribute values} 
      \label{fig:interp_music}
    \end{figure*}

    \paragraph{Music:} Fig.~\ref{fig:interp_music} shows the results of manipulating different musical attributes in the Bach Chorales and Folk Music datasets. For AR-VAE, in most cases, traversal along the regularized dimensions lead to measures with increasing values of the attributes. These can be seen more clearly in the attribute value plots to the right of the piano-rolls. 
    
    On the contrary, for $\beta$-VAE, the progression of the attributes is not as uniform. In fact, in the Bach Chorales example, there is no change in the values of two out of the four attributes. This is because, in the case of $\beta$-VAE, on many occasions there is a single dimension of the latent space which has maximum mutual information with two or more attributes. This could be either due to a high degree of correlation between attributes (e.g., \textit{Rhythmic Complexity} and \textit{Note Density}) or due to very poor disentanglement where all attributes are poorly encoded. Even in cases where the attributes have a nice progression (e.g., see bottom right row for the \textit{Contour} attribute in the Folk Music examples in Fig.~\ref{fig:interp_music}), the order of the encoding is reversed (the attribute decreases with increasing value of the latent code). This warrants a post-hoc analysis of the latent space to understand the relationship between the attributes and the latent dimensions.

    It is important to point out that while AR-VAE allows better control over the different attributes for both datasets, it sometimes struggles to maintain the musical coherency of the generated measures. For instance, the generated measures shown in Fig.~\ref{fig:interp_music} are not always in the same key. Although this seems to the case for measures generated with the $\beta$-VAE model as well, the problem seems to be more pronounced for AR-VAE. We observed that training separate models for the different attributes, where each model regularizes a single attribute, tends to alleviate this limitation. Traversing along the regularized dimensions in these models typically results in better musical coherence.

  \subsection{Latent Space Visualization}
  \label{exp_interpretability}

    \begin{figure}[t]
      \centering
      \begin{tabular}{@{}c@{}}
        \includegraphics[width=0.23\textwidth]{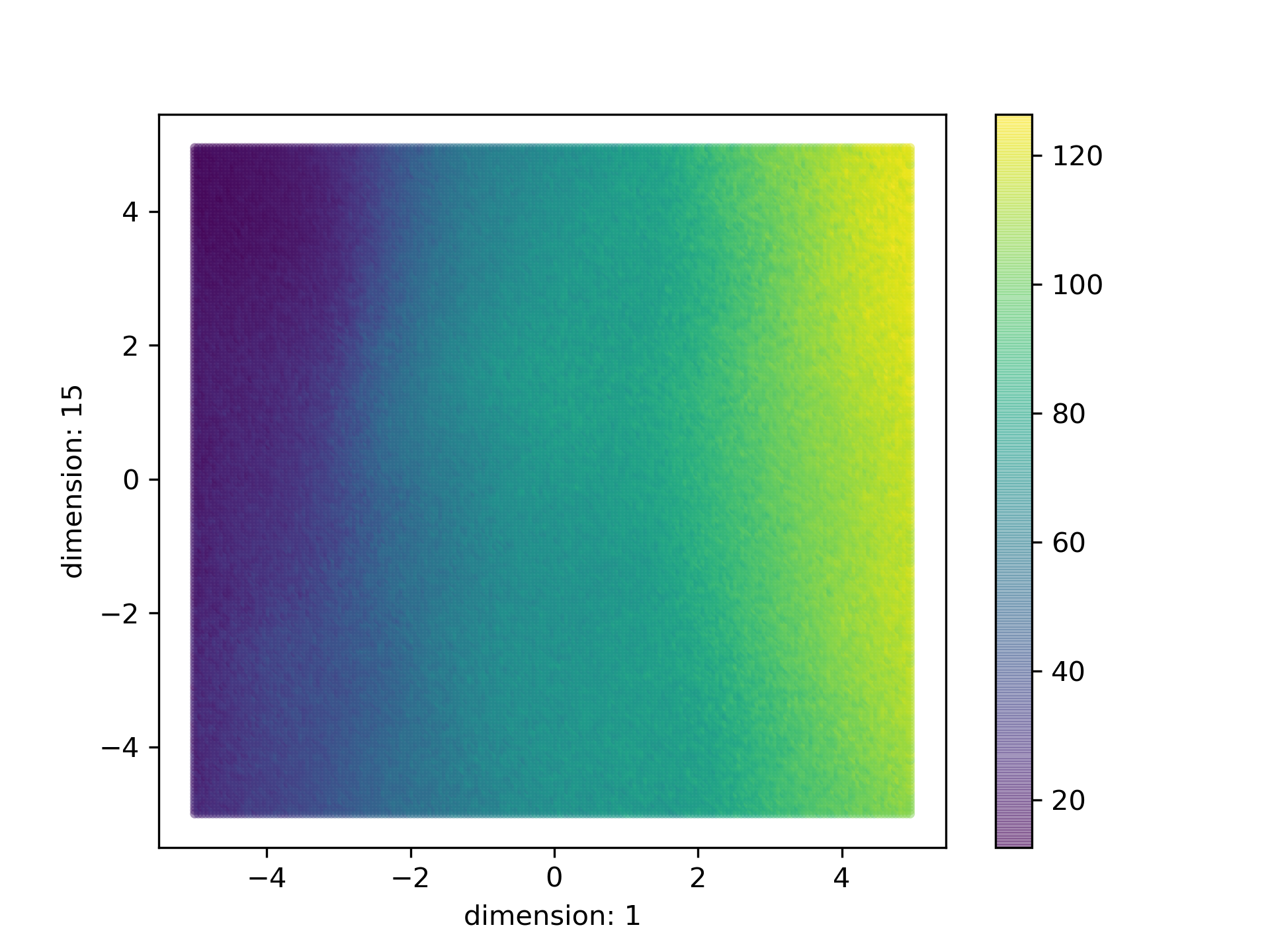} \\[\abovecaptionskip]
        \small (a) \textit{Area}
      \end{tabular}
      \begin{tabular}{@{}c@{}}
        \includegraphics[width=0.23\textwidth]{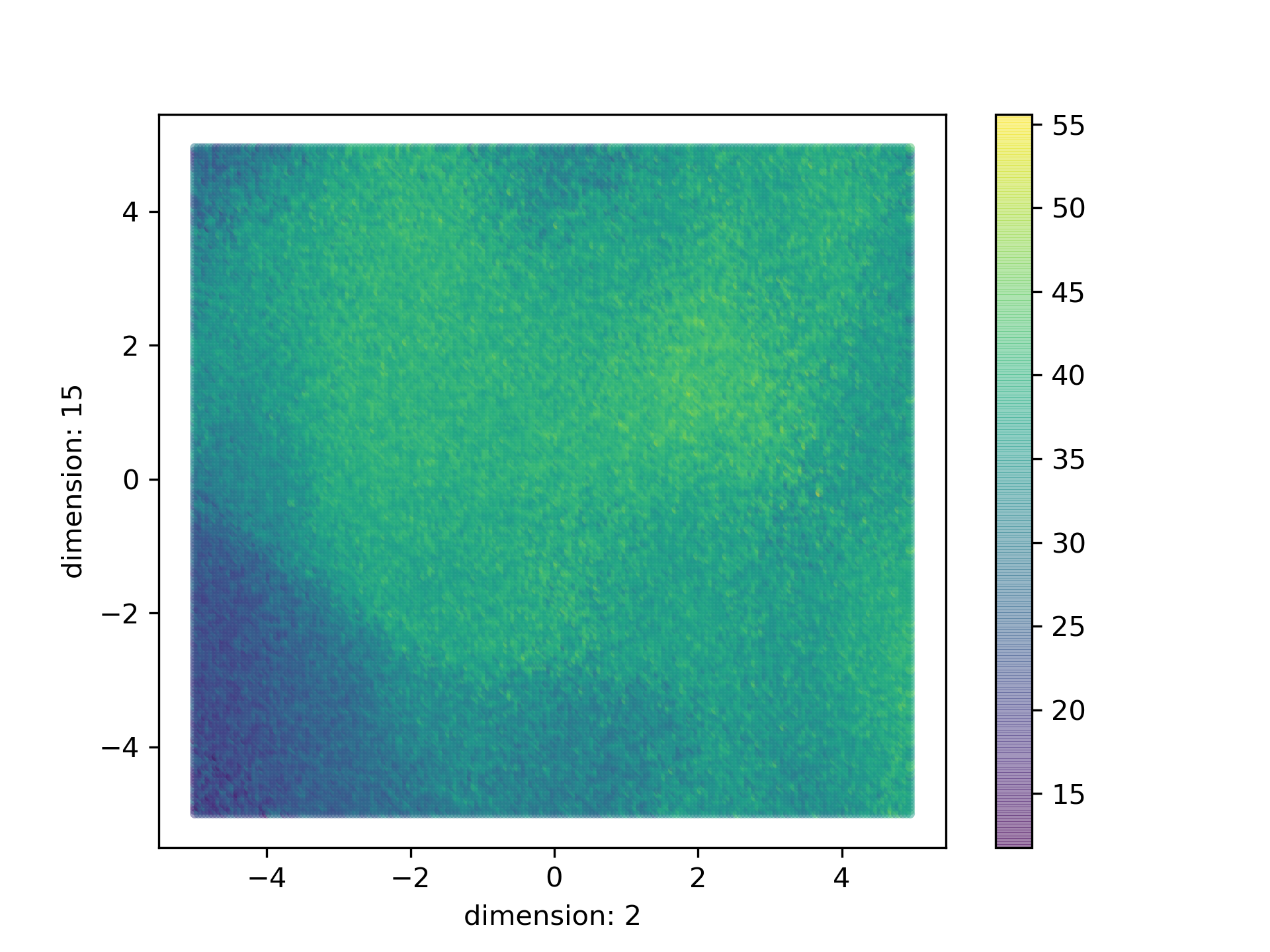} \\[\abovecaptionskip]
        \small (b) \textit{Length}
      \end{tabular}
      \begin{tabular}{@{}c@{}}
        \includegraphics[width=0.23\textwidth]{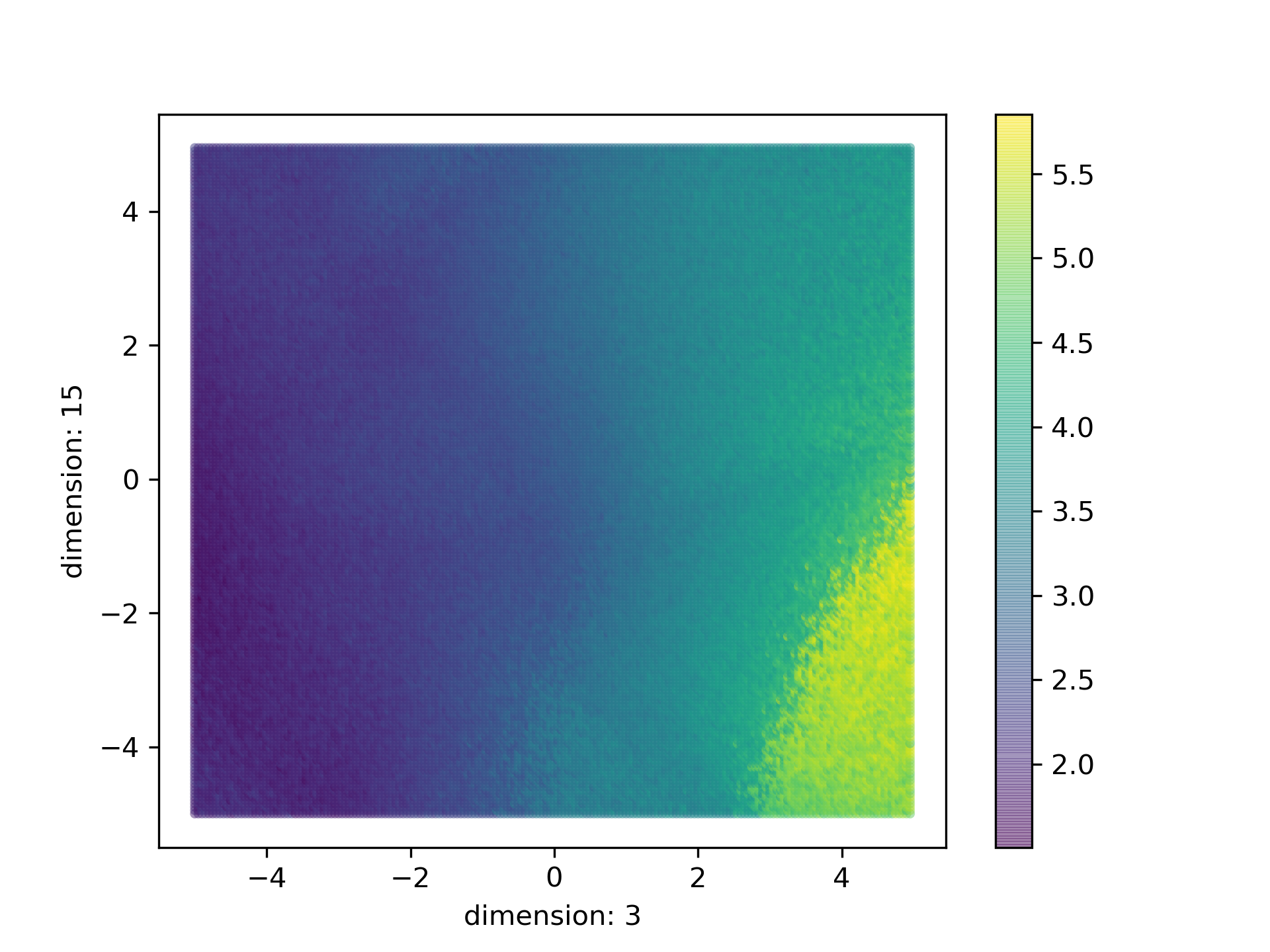} \\[\abovecaptionskip]
        \small (c) \textit{Thickness}
      \end{tabular}
      \begin{tabular}{@{}c@{}}
        \includegraphics[width=0.23\textwidth]{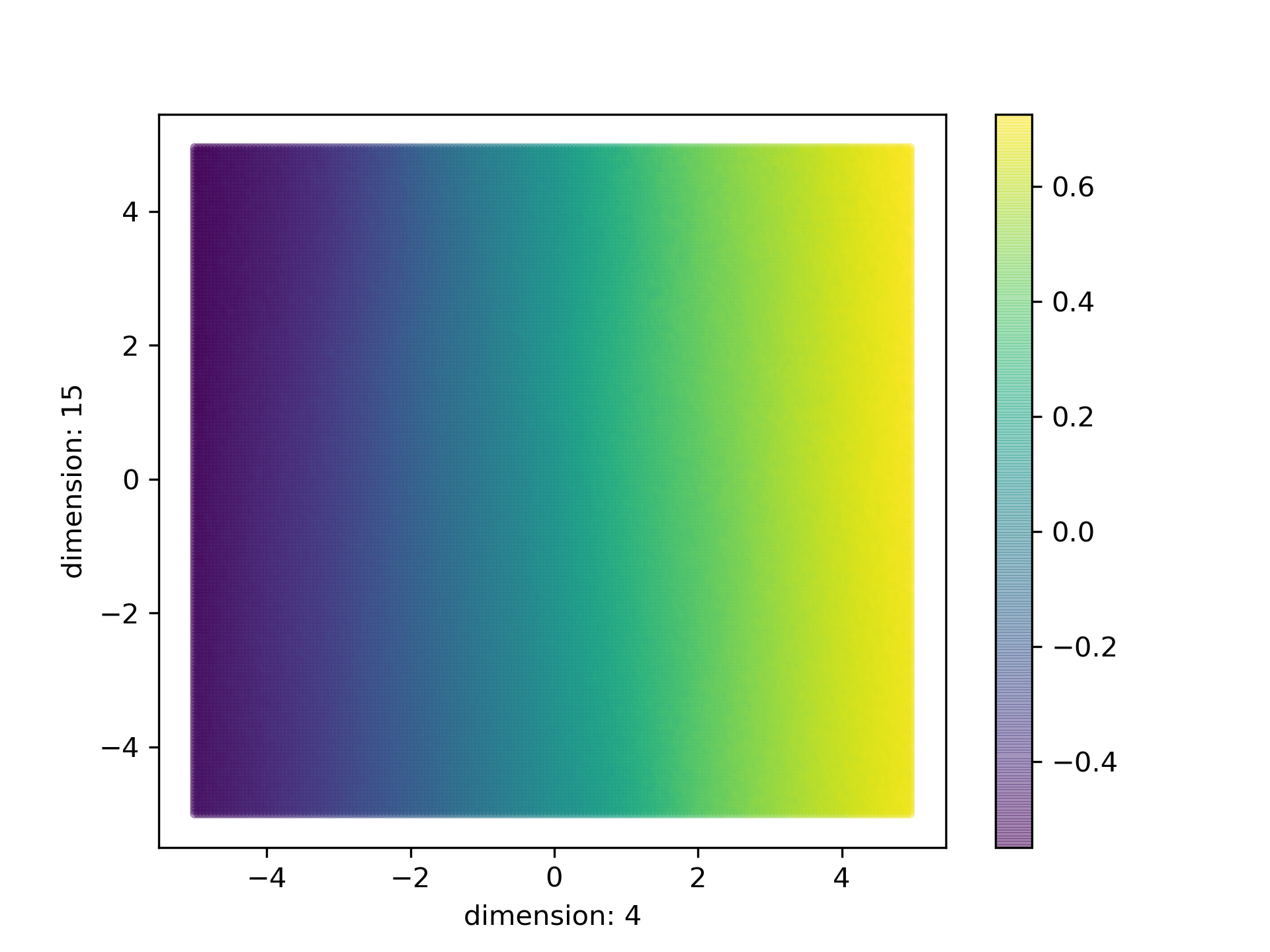} \\[\abovecaptionskip]
        \small (d) \textit{Slant}
      \end{tabular}
      \begin{tabular}{@{}c@{}}
        \includegraphics[width=0.23\textwidth]{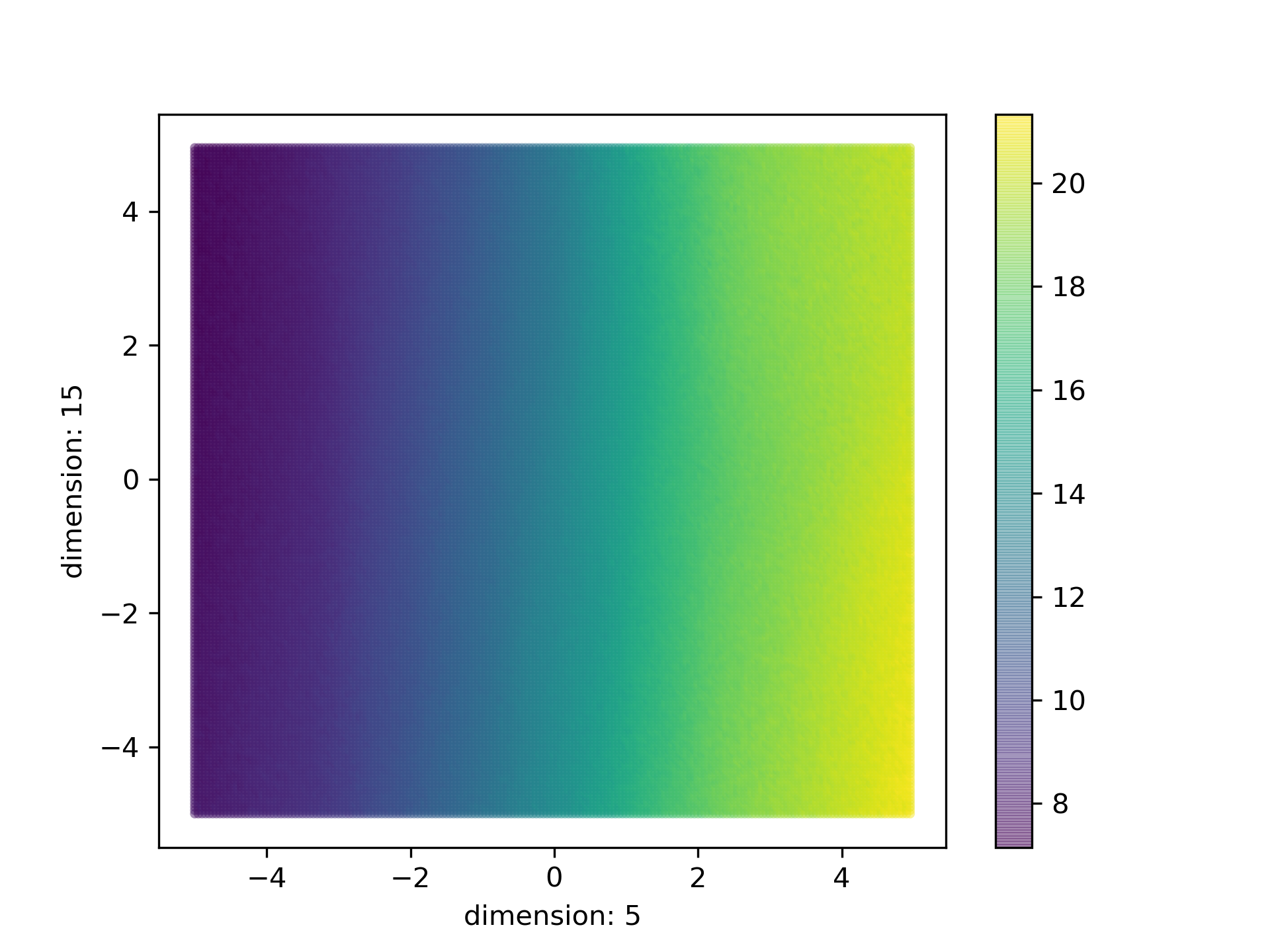} \\[\abovecaptionskip]
        \small (e) \textit{Width}
      \end{tabular}
      \begin{tabular}{@{}c@{}}
        \includegraphics[width=0.23\textwidth]{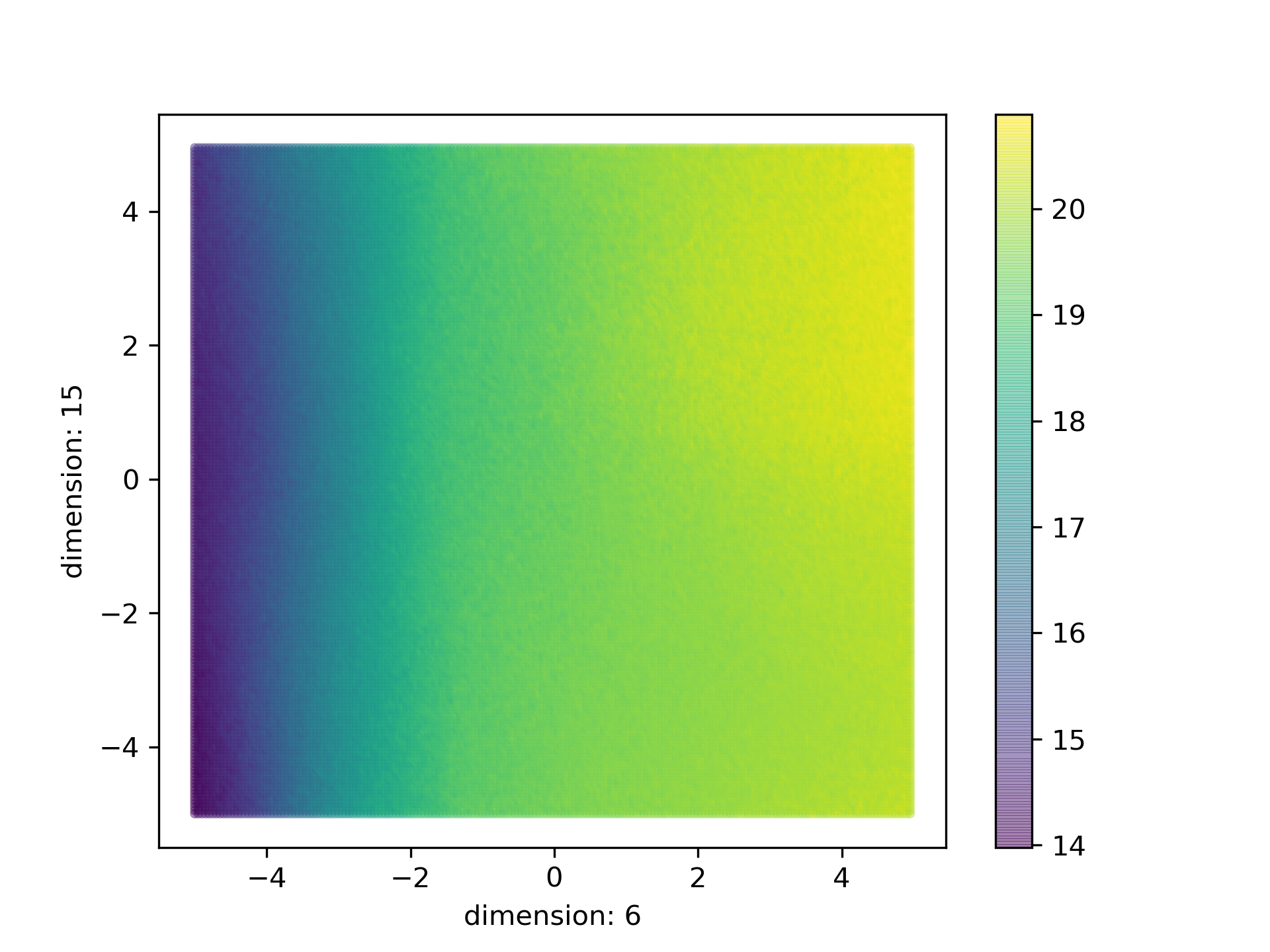} \\[\abovecaptionskip]
        \small (f) \textit{Height}
      \end{tabular}
      \caption{Attribute surface plots for the Morpho-MNIST dataset. Each plots shows the attribute values for the decoded latent vectors on a 2-dimensional surface of the latent space (the latent code for the other dimensions is kept fixed). The \textit{x}-axis corresponds to the regularized dimension for the attribute and the \textit{y}-axis corresponds to a non-regularized dimension. The attribute values increase (gradual transition from purple to yellow color) as the latent code for the regularized dimension is increased for most attributes} 
      \label{fig:lsv_mnist}
    \end{figure}

    \begin{figure}[t]
      \centering
      \begin{tabular}{@{}c@{}}
        \includegraphics[width=0.23\textwidth]{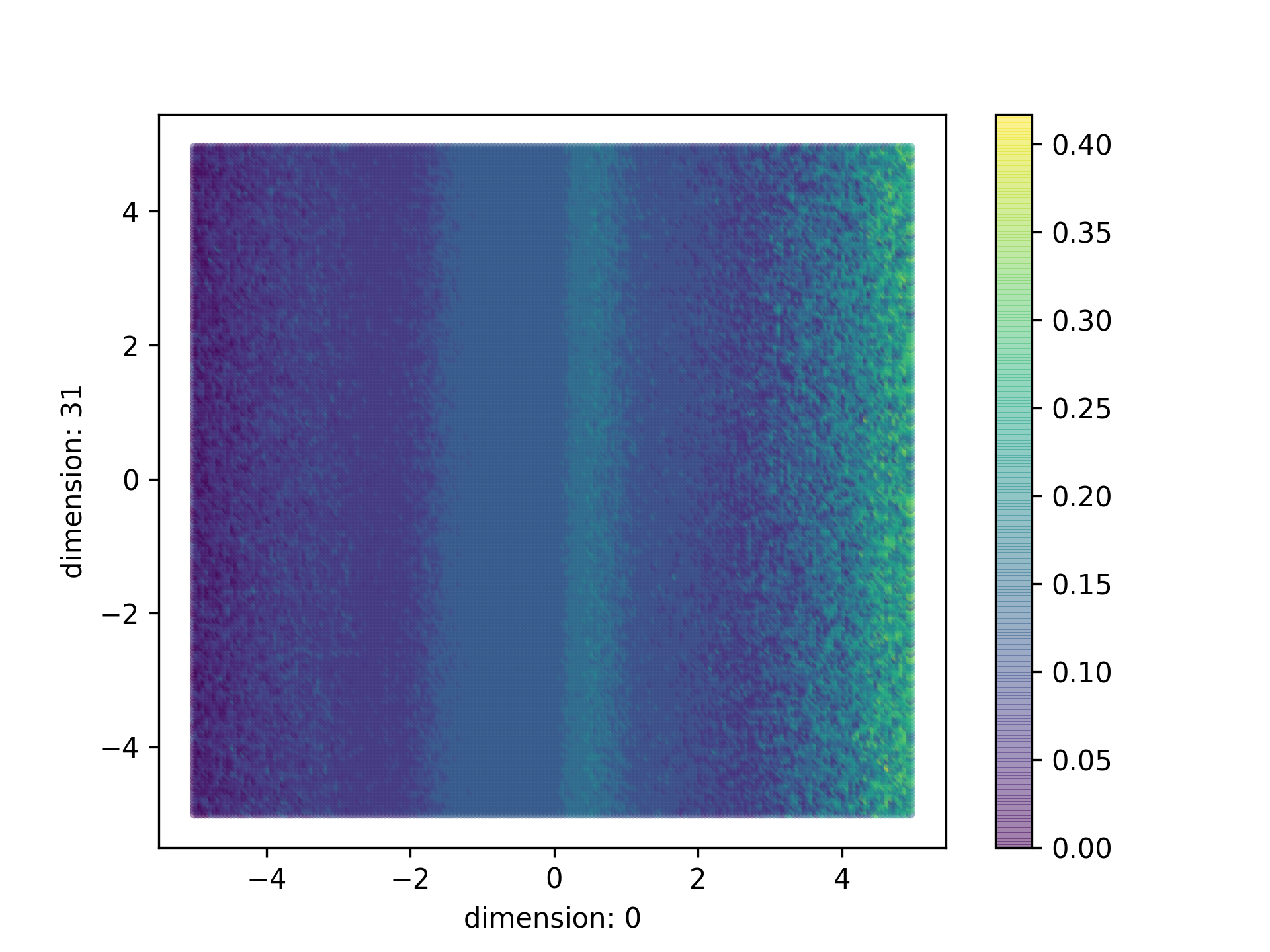} \\[\abovecaptionskip]
        \small (a) \textit{Rhythmic Complexity}
      \end{tabular}
      \begin{tabular}{@{}c@{}}
        \includegraphics[width=0.23\textwidth]{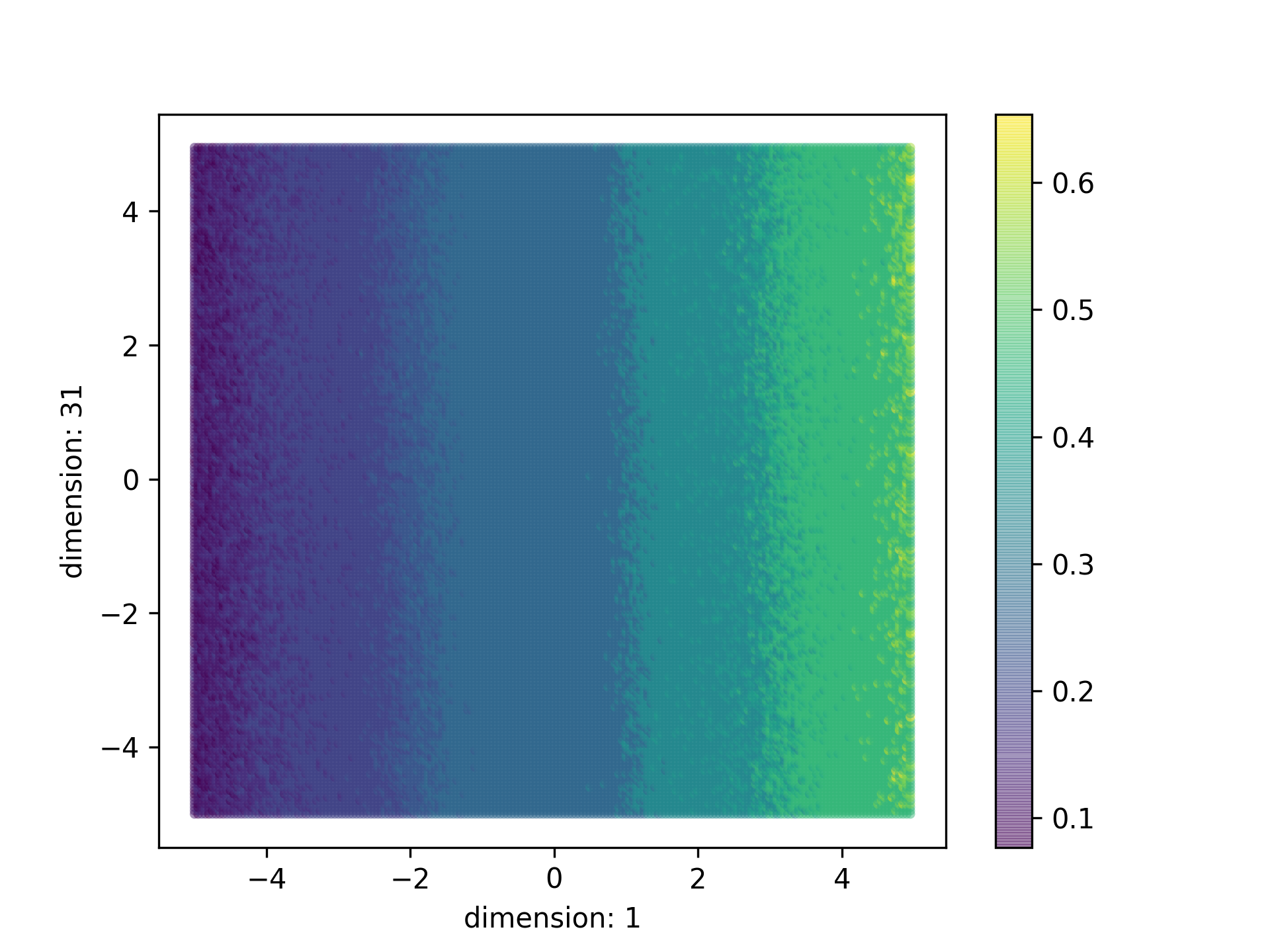} \\[\abovecaptionskip]
        \small (b) \textit{Pitch Range}
      \end{tabular}
      \begin{tabular}{@{}c@{}}
        \includegraphics[width=0.23\textwidth]{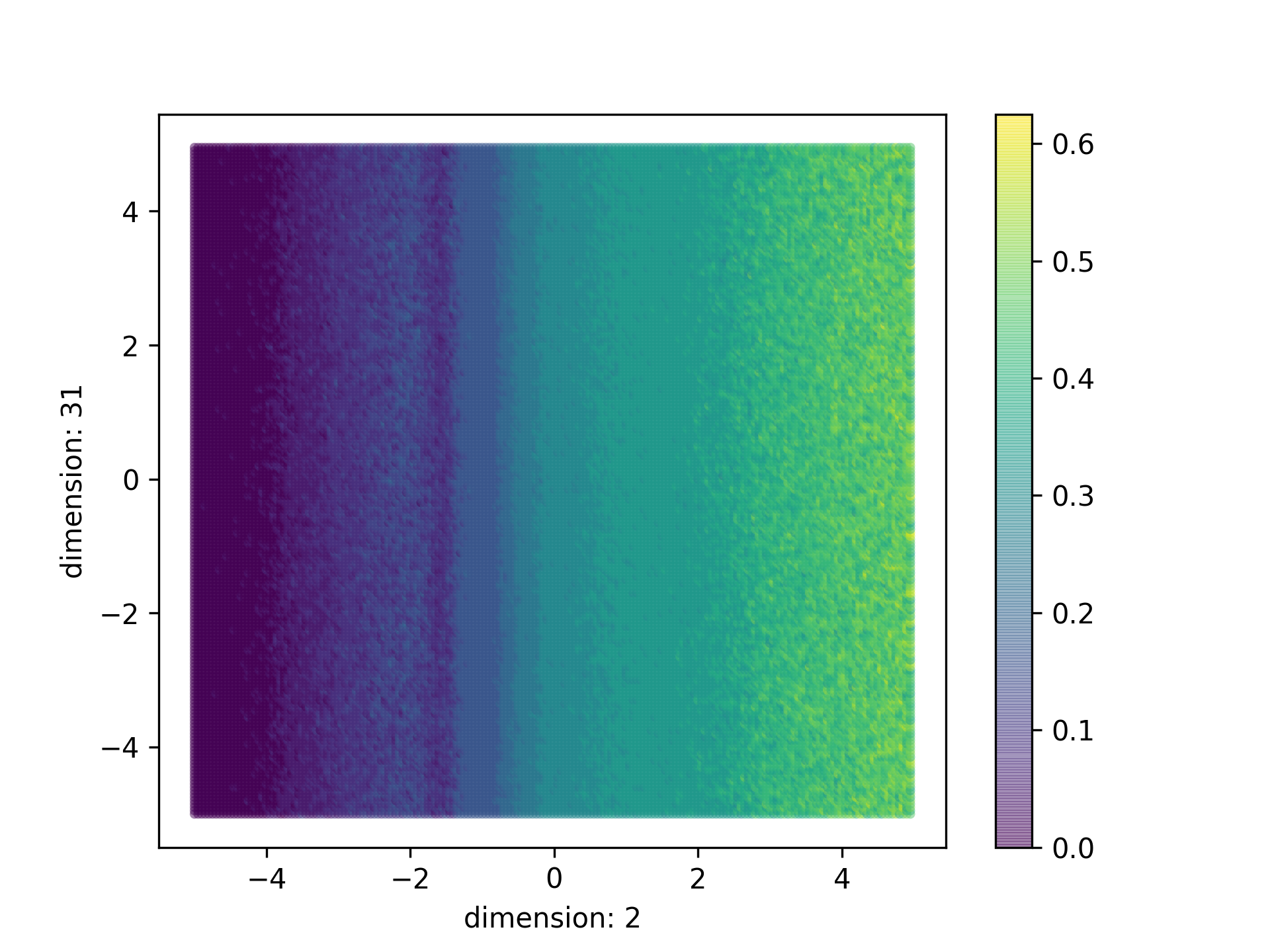} \\[\abovecaptionskip]
        \small (c) \textit{Note Density}
      \end{tabular}
      \begin{tabular}{@{}c@{}}
        \includegraphics[width=0.23\textwidth]{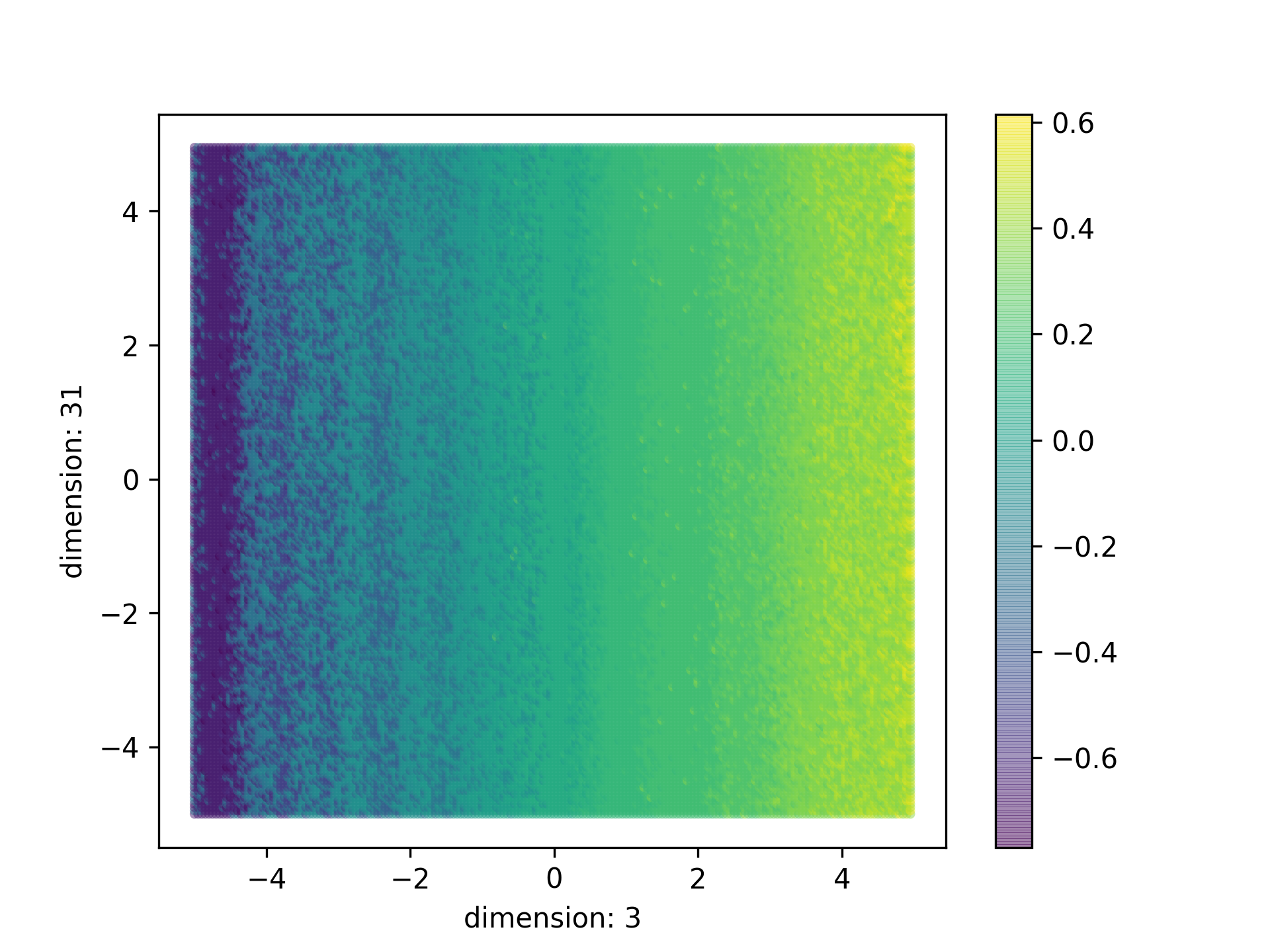} \\[\abovecaptionskip]
        \small (d) \textit{Contour}
      \end{tabular}
      \caption{Attribute surface plots for the Folk Music dataset. Each plot shows the attribute values for the decoded latent vectors on a 2-dimensional surface of the latent space (the latent code for the other dimensions is kept fixed). The \textit{x}-axis corresponds to the regularized dimension for the attribute and the \textit{y}-axis corresponds to a non-regularized dimension. The attribute values increase (gradual transition from purple to yellow color) as the latent code for the regularized dimension is increased for most attributes} 
      \label{fig:lsv_folk}
    \end{figure}
    
    In this experiment, the interpretability of the latent space with respect to different attributes is investigated by generating attribute surface plots. For each attribute, A 2-dimensional plane on the latent space is considered which comprises of the regularized dimension (\textit{x}-axis) for the attribute and a non-regularized dimension (\textit{y}-axis). The latent code for the other dimensions are drawn from a normal distribution and kept fixed. The latent vectors thus obtained are passed through the decoder and the attributes of the generated data are computed. Fig.~\ref{fig:lsv_mnist} and \ref{fig:lsv_folk} show the result of this visualization for the Morpho-MNIST and the Folk Music dataset, respectively. 

    For both datasets, most attributes show an increasing trend as the latent code of the regularized dimension is increased. This is seen by the gradual transition from purple to yellow color along the \textit{x}-axis. The change in color is minimal for the \textit{y}-axis which indicates independence of the attribute from the non-regularized dimension. Thus, the AR-VAE latent space is interpretable with respect to the attribute values and traversing along the regularized dimension will lead to an increase in the corresponding attribute. The only attributes for which this does not hold are \textit{Length} and \textit{Rhythmic Complexity} in the morpho-MNIST and Folk Music datasets, respectively. The reason for the poor performance for \textit{Rhythmic Complexity} could be that it is strongly correlated with \textit{Note Density} ($\approx 0.89$). This is also reflected in the relatively poor \textit{Interpretability} score for \textit{Rhythmic Complexity} ($0.83$) compared to the other attributes ($\approx 0.99$). The choice of attributes and their relationship to each other, thus, becomes an important consideration to create meaningful latent spaces.

\section{Conclusion}
\label{sec:conclusion}
This paper investigates the problem of selective manipulation of data attributes in deep generative models by structuring the latent space of a VAE to encode specific attributes along specific dimensions of the latent space. The proposed AR-VAE model uses a novel regularization loss to enforce a monotonic relationship between the attributes and the latent code of the respective regularized dimension. The resulting latent spaces are easily interpretable and allow manipulation of individual attributes by simple traversals along the regularized dimensions. The regularization loss works for continuous data attributes, and has a simple formulation with only two hyperparameters. In addition, contrary to previous supervised methods \cite{hadjeres_glsr-vae_2017,kulkarni_deep_2015}, the loss formulation is agnostic to how the attributes are computed or obtained. Using both image and music-based datasets, we show that AR-VAE can work with different types of data, model architectures, and a wide range of data attributes. We show that AR-VAE leads to better disentanglement of the latent space, and creates meaningful attribute-based interpolations while preserving the content of the original data. This superior performance is achieved without any significant drop in the reconstruction quality of the VAE. \ashis{While AR-VAE is designed to work with VAE-based architectures, the proposed regularization method should also work for other types of generative models such as Auto-Encoders \cite{vincent_extracting_2008}, GANs \cite{goodfellow_generative_2014}, and flow-based models \cite{rezende_variational_2015}. }

\ashis{The proposed method can be used as a building block to create several interesting and useful applications for context-driven data generation. For instance, it can be used to manipulate attributes in image or photo-based applications such as FaceApp\footnote{https://faceapp.com/app, last accessed: 20th July 2020} or Prisma.\footnote{https://prisma-ai.com, last accessed: 20th July 2020} In the context of music, it can be used to build interactive tools or plugins to aid composers and music creators. For instance, composers would be able to manipulate different attributes (such as rhythmic complexity) of the composed music to try different ideas and meet specific compositional requirements. This would allow fast iteration and would be especially useful for novices and hobbyists. Since the method is agnostic to how the attributes are computed, it can potentially be useful to manipulate high-level musical attributes such as tension and emotion. This will be particularly useful for music generation in the context of video games where the background music can be suitably changed to match the emotional context of the game and the actions of the players. There is also the possibility of using this for speech generation applications. For instance, the ability to manipulate the prosody of the generated speech can make mobile voice assistants more realistic.}

There are, however, some limitations of the approach which can open up avenues for future research. The regularization loss is currently designed to work with continuous attributes. While we present evidence that it also works for discrete categorical attributes (e.g., \textit{shape} in the 2-d sprites dataset), additional experiments will be needed to ascertain this. Furthermore, the current formulation is not suitable for binary attributes such as the ones used in Fader networks \cite{lample_fader_2017}. We also observe that the choice of attributes seems to play an important role in the training process. While we are able to jointly regularize multiple attributes in most cases (as seen in Sect.~\ref{exp_manipulation} and \ref{exp_interpretability}), strongly correlated attributes can lead to latent spaces which are not interpretable with respect to every single attribute. This results in poor control over some attributes and reduced content preservation and coherence in the interpolations. While this is a limitation, we think that independent control over strongly correlated attributes is probably not a strict requirement for a useful generative model. Even in cases where such control is necessary, two or more separately trained AR-VAE models can be used which regularize those attributes individually. 

\section*{Acknowledgements}
  \ashis{The authors would like to thank Nvidia Corporation for their donation of a Titan V awarded as part of the GPU (Graphics Processing Unit) grant program which was used for running several experiments pertaining to this research work.}

\section*{Conflict of interest}
  The authors declare that they have no conflict of interest.

\bibliographystyle{spmpsci}      
\bibliography{ar-vae}   

\clearpage

\appendix
  \section*{APPENDIX}

  \section{Computation of Musical Attributes}
  \label{app:musical-metrics}  
  The data representation scheme from \cite{pati_learning_2019} is chosen where each monophonic measure of music $M$ is a sequence of $N$ symbols $\left\{ m_t \right\}, t \in [0, N)$, where $N=24$. The set of symbols consists of note names (e.g., A\#, Eb, B, C), a continuation symbol `\_\_', and a special token for Rest. The computation steps for the musical metrics are as follows:
  \begin{compactenum}[(a)]
    \item \textit{Rhythmic Complexity (r)}: This attribute measures the rhythmic complexity of the a given measure. To compute this, a complexity coefficient array $\left\{ f_t \right\}, t \in [0, N)$ is first constructed which assigns weights to different metrical locations based on Toussaint’s metrical complexity measure \cite{toussaint_mathematical_2002}. Metrical locations which are on the beat are given low weights while locations which are off-beat are given higher weights. The attribute is computed by taking a weighted average of the note onset locations with the complexity coefficient array $f$. Mathematically, 
      \begin{equation}
        \label{eq:rhy_complexity}
        r(M) = \frac{\sum_{t=0}^{N-1} \mathrm{ONSET}(m_t) . f_t}{\sum_{t=0}^{N-1}f_t} ,
      \end{equation}
    where $\mathrm{ONSET} \left ( \cdot \right )$ detects if there is a note onset at location $t$, i.e., it is $1$ if $m_t$ is a note name symbol and $0$ otherwise.
    
    \item \textit{Pitch Range (p)}: This is computed as the normalized difference between the maximum and minimum MIDI pitch values:
      \begin{equation}
        \label{eq:pitch_range}
        p(M) = \frac{1}{R} \left( \underset{t \in [0, N)}{\mathrm{max}}(\mathrm{MIDI}(m_t)) - \underset{t \in [0, N)}{\mathrm{min}}(\mathrm{MIDI}(m_t)) \right),
      \end{equation}
    where $\mathrm{MIDI} \left ( \cdot \right )$ computes the pitch value in MIDI for the note symbol. The MIDI pitch value for Rest and `\_\_' symbols are set to zero. The normalization factor $R$ is based on the range of the dataset. 
    
    \item \textit{Note Density (d)}: This measures the count of the number of notes per measure normalized by the total length of the measure sequence:
      \begin{equation}
        \label{eq:note_density}
        d(M) = \frac{1}{N} \sum_{i=0}^{N-1} \mathrm{ONSET}(m_t),
      \end{equation}
    where $\mathrm{ONSET} \left ( \cdot \right )$ has the same meaning as in Eq.~(\ref{eq:rhy_complexity}) above.
    
    \item \textit{Contour (c)}: This measures the degree to which the melody moves up or down and is measured by summing up the difference in pitch values of all the notes in the measure. Mathematically, 
      \begin{equation}
        \label{eq:contour}
        c(M) = \frac{1}{R} \sum_{t=0}^{N-2} \left[ \mathrm{MIDI}(m_{t+1}) - \mathrm{MIDI}(m_t) \right],
      \end{equation}
    where $\mathrm{MIDI} \left ( \cdot \right )$ and $R$ have same meaning as in Eq.~(\ref{eq:pitch_range}) above.
  \end{compactenum}

  \section{Implementation Details}
  \label{app:model-arch}    
    \paragraph{Image-based Models:} 
      For the image-based models, a stacked convolutional VAE architecture is used. The encoder consists of a stack of $N$ 2-dimensional convolutional layers followed by a stack of linear layers. The decoder mirrors the encoder and consists of a stack of linear layers followed by a stack of $N$ 2-dimensional transposed convolutional layers. The configuration details are given in Table~\ref{tab:image_model_config}.

\begin{figure}[h]
        \centerline{
        \includegraphics[width=\columnwidth]{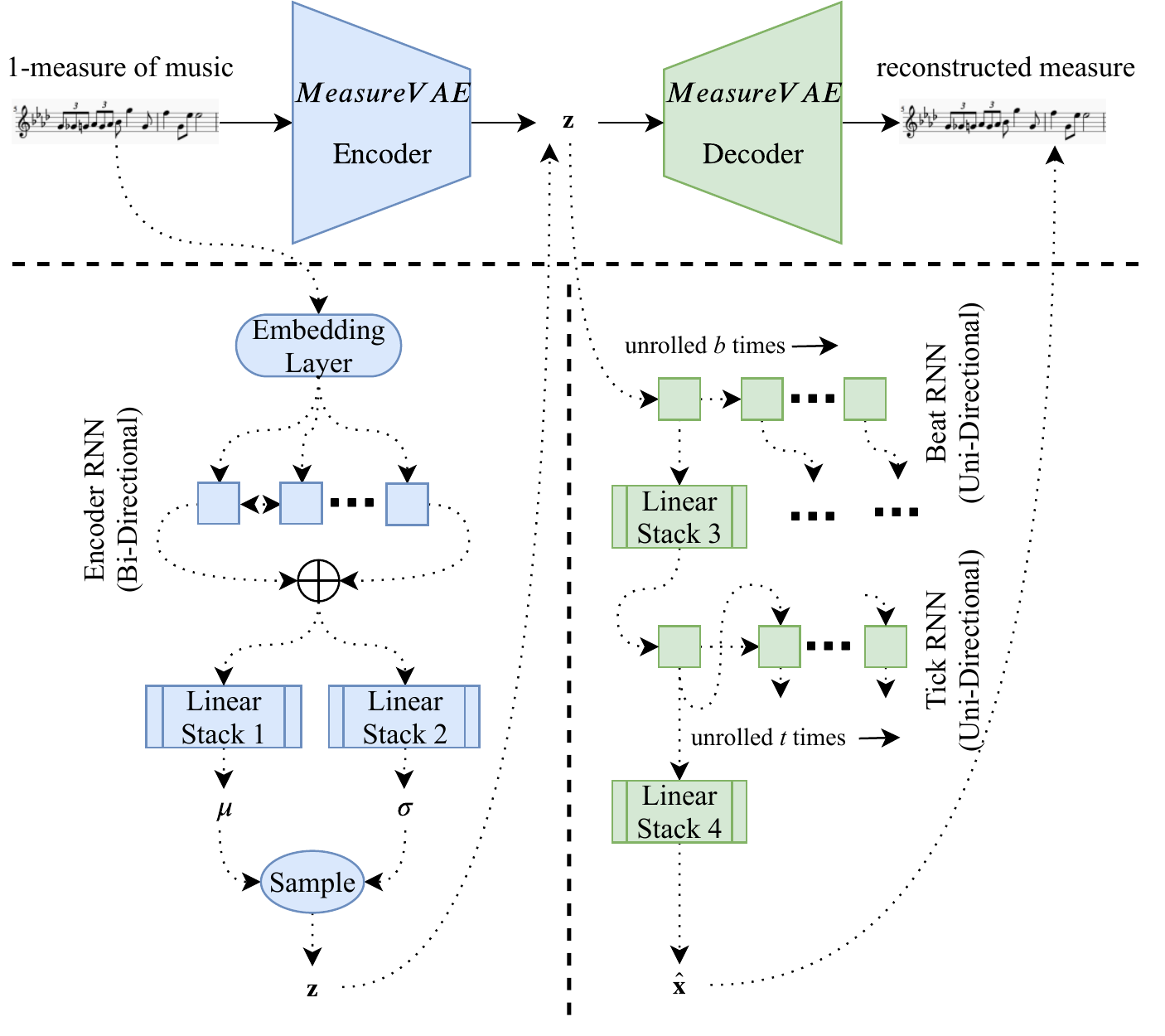}}
        \caption{MeasureVAE schematic. Individual components of the encoder and decoder are shown below the main blocks (dotted arrows indicate data flow within the individual components). $\mathbf{z}$ denotes the latent vector and $\hat{\mathbf{x}}$ denotes the reconstructed measure, $b=4$ denotes the number of beats in a measure and $t=6$ denotes the number of symbols/ticks in a beat. Fig. taken from \cite{pati_learning_2019}}
      \label{fig:measurevae_schematic}
      \end{figure}

    \paragraph{Music-based Models:}
      For the music-based models, the model architecture is based on our previous work of musical score inpainting \cite{pati_learning_2019}. A hierarchical recurrent VAE architecture is used. Fig.~\ref{fig:measurevae_schematic} shows the overall schematic of the architecture and Table~\ref{tab:music_model_config} provides the configuration details. 

      \begin{table}[h]
        \footnotesize
        \begin{center}
        \begin{tabularx}{\columnwidth}{Xl}
            \toprule
            \multicolumn{2}{c}{\textit{Measure VAE}} \\ 
            \toprule
            Embedding Layer  & i=dict size, o=10   \\ \midrule
            EncoderRNN       & n=2, i=10, h=128, d=0.5 , type=GRU \\ \midrule
            \begin{tabular}[c]{@{}l@{}}Linear Stack 1 \\ Linear Stack 2\end{tabular} & i=512, o=32, n=2, non-linearity=SELU \\ \midrule
            BeatRNN          & n=2, i=1, h=128, d=0.5, type=GRU   \\ \midrule
            TickRNN          & n=2, i=138, h=128, d=0.5, type=GRU  \\ \midrule
            Linear Stack 3   & i=128, o=256, n=1, non-linearity=SELU \\ \midrule
            Linear Stack 4   & i=128, o=dict size, n=1, non-linearity=ReLU \\ \bottomrule
        \end{tabularx}
        \end{center}
      \caption{Table showing configurations of the MeasureVAE architecture. n: number of layers, i: input size, o: output size, h: hidden size, d: dropout probability, SELU: Scaled Exponential Linear Unit \cite{klambauer_self-normalizing_2017}, ReLU: Rectifier Linear Unit, GRU: Gated Recurrent Units \cite{jozefowicz_empirical_2015}}
      \label{tab:music_model_config}
      \end{table}

    \paragraph{Training Details:}
      All models for the same dataset are trained for the same number of epochs (models for both image-based datasets and Bach Chorales are trained for $100$ epochs, models for the Folk Music dataset are trained for $30$ epochs). The optimization is carried out using the ADAM optimizer \cite{kingma_adam_2015} with a fixed learning rate of $1\mathrm{e}{-4}$, $\beta _1 = 0.9$, $\beta _2 = 0.999$, and $\epsilon=1$e$-8$. 

      \begin{table*}[t]
        \footnotesize
        \begin{center}
        \begin{tabularx}{\textwidth}{Xll}
          \toprule
          Model Type & 2-d Sprites VAE & Mnist VAE\\ \toprule
            \begin{tabular}[c]{@{}l@{}}
              Encoder Convolutional  \\ 
              Stack
            \end{tabular} 
            & \begin{tabular}[c]{@{}l@{}}
                4-layer Convolutional Network: \\
                Conv(i=1, o=32, k=4, s=2, p=1) + ReLU \\ 
                Conv(i=32, o=32, k=4, s=2, p=1) + ReLU \\ 
                Conv(i=32, o=32, k=4, s=2, p=1) + ReLU \\
                Conv(i=32, o=32, k=4, s=2, p=1) + ReLU \\
              \end{tabular} 
            & \begin{tabular}[c]{@{}l@{}}
                3-layer Convolutional Network: \\
                Conv(i=1, o=64, k=4, s=1, p=0) + SELU + Dropout(d=0.5) \\ 
                Conv(i=64, o=64, k=4, s=1, p=0) + SELU + Dropout(d=0.5) \\
                Conv(i=64, o=8, k=4, s=1, p=0) + SELU + Dropout(d=0.5) \\
              \end{tabular}  \\ \midrule
            \begin{tabular}[c]{@{}l@{}}
              Encoder Linear  \\ 
              Stack
            \end{tabular} 
            & \begin{tabular}[c]{@{}l@{}}
                3-layer Linear Network: \\
                Linear(i=512, o=256) + ReLU \\
                Linear(i=256, o=256) + ReLU \\
                Linear(i=256, o=10) $\times$ 2 (in parallel)\\
              \end{tabular} 
            & \begin{tabular}[c]{@{}l@{}}
                2-layer Linear Network: \\
                Linear(i=2888, o=256) + SELU \\
                Linear(i=256, o=16) $\times$ 2 (in parallel) \\
            \end{tabular}  \\ \midrule
          \begin{tabular}[c]{@{}l@{}}
            Decoder Linear  \\ 
            Stack
          \end{tabular}
            & \begin{tabular}[c]{@{}l@{}}
                3-layer Linear Network: \\
                Linear(i=10, o=256) + ReLU \\
                Linear(i=256, o=256) + ReLU \\
                Linear(i=256, o=512) + ReLU \\
              \end{tabular} 
            & \begin{tabular}[c]{@{}l@{}}
                2-layer Linear Network: \\
                Linear(i=16, o=256) + SELU \\
                Linear(i=256, o=2888) + SELU \\
              \end{tabular}  \\ \midrule
            \begin{tabular}[c]{@{}l@{}}
              Decoder Convolutional  \\ 
              Stack
            \end{tabular}
            & \begin{tabular}[c]{@{}l@{}}
                4-layer Transposed Convolutional Network: \\
                TrConv(i=32, o=32, k=4, s=2, p=1) + ReLU \\ 
                TrConv(i=32, o=32, k=4, s=2, p=1) + ReLU \\ 
                TrConv(i=32, o=32, k=4, s=2, p=1) + ReLU \\
                TrConv(i=32, o=1, k=4, s=2, p=1) \\
              \end{tabular} 
            & \begin{tabular}[c]{@{}l@{}}
                3-layer Transposed Convolutional Network: \\
                TrConv(i=8, o=64, k=4, s=1, p=0) + SELU + Dropout(d=0.5) \\ 
                TrConv(i=64, o=64, k=4, s=1, p=0) + SELU + Dropout(d=0.5) \\
                TrConv(i=64, o=1, k=4, s=1, p=0) \\
              \end{tabular}  \\ \bottomrule
        \end{tabularx}
        \end{center}
      \caption{Table showing configurations of the VAEs for the image-based datasets.In the Encoder Linear Stack, the last layer has two parallel linear layers for computing the mean and log standard deviation of the latent vectors respectively.  Conv: 2-dimensional convolutional layer, TrConv: 2-dimensional transposed convolutional layer, i: input channels, o: output channels, k: kernel size, s: stride, p: padding, d: dropout probability, SELU: Scaled Exponential Linear Unit \cite{klambauer_self-normalizing_2017}, ReLU: Rectifier Linear Unit}
      \label{tab:image_model_config}
      \end{table*} 

      \ashis{All the models are implemented using the Python programming language and the Pytorch\footnote{https://pytorch.org, last accessed: 20th July 2020} library.} 

  \section{Additional Results}
  \label{app:add_results}
\begin{figure*}[t]
      \centerline{
      \includegraphics[width=0.9\textwidth]{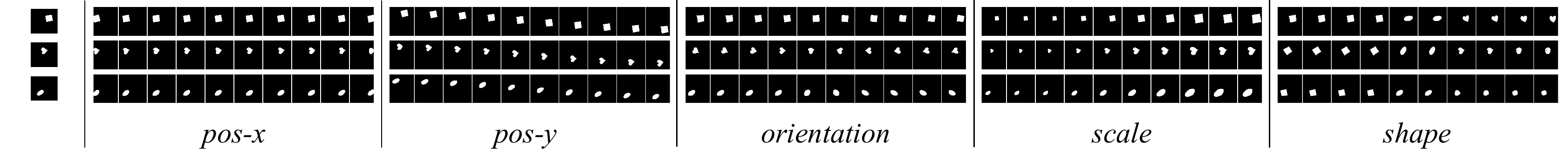}}
      \caption{Manipulating attributes for three different shapes from the 2-d sprites dataset images using AR-VAE. The interpolations for each attribute are generated by changing the latent code of the corresponding regularized dimension for the original shapes shown on the extreme left. Attributes can be manipulated independently}
    \label{fig:dsprites_add_results}
    \end{figure*}

\begin{figure*}[t]
      \centerline{
      \includegraphics[width=\textwidth]{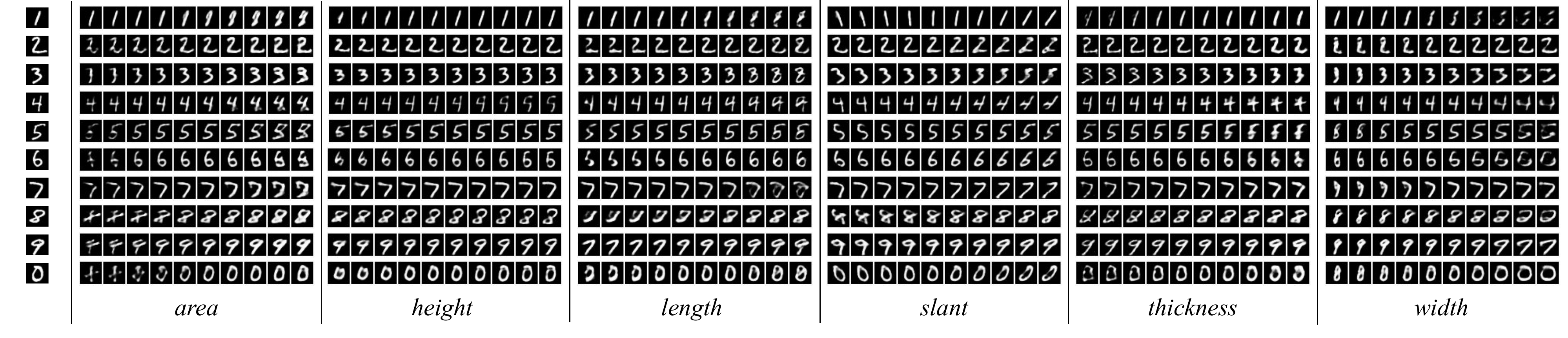}}
      \caption{Manipulating attributes for ten different digits from the Morpho-MNIST dataset images using AR-VAE. The interpolations for each attribute are generated by changing the latent code of the corresponding regularized dimension for the original digits shown on the extreme left. AR-VAE is able to manipulate the different attributes and is able to retain the digit identity in most cases}
    \label{fig:mnist_add_results}
    \end{figure*}

\begin{figure*}[t]
      \centerline{
      \includegraphics[width=\textwidth]{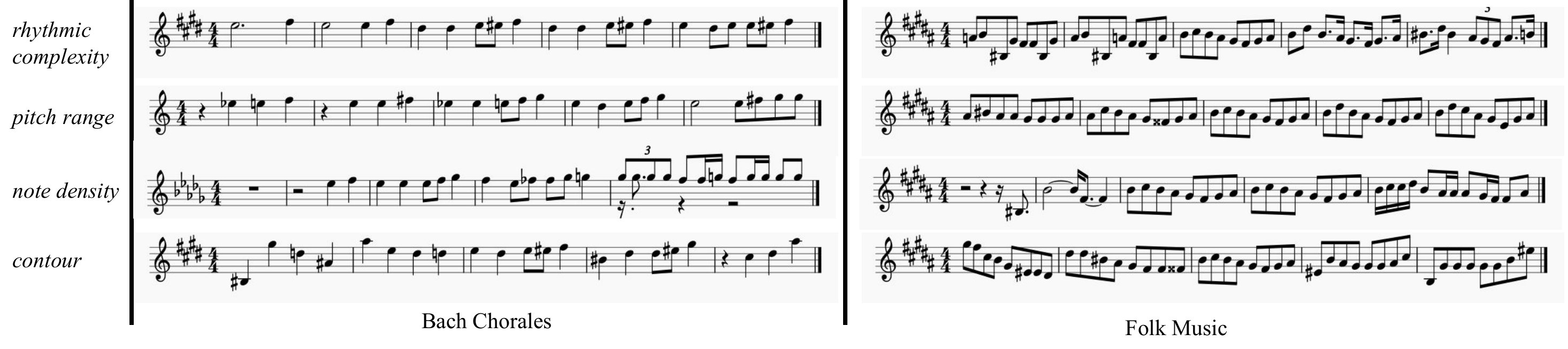}}
      \caption{Musical score corresponding to the AR-VAE generated interpolations from Fig.~\ref{fig:interp_music}. While the attribute values of the generated measures are controlled effectively, the musical coherence is often lost (particularly in the case of Bach Chorales)}
    \label{fig:interp_score}
    \end{figure*}

    Some additional examples from the image-based datasets are shown in Fig.~\ref{fig:dsprites_add_results} and \ref{fig:mnist_add_results}. The musical scores for AR-VAE generated interpolations from Fig.~\ref{fig:interp_music} is shown in Fig.~\ref{fig:interp_score}. 

\end{document}